\pdfoutput=1

\documentclass[11pt]{article}

\usepackage[]{acl}

\usepackage{times}
\usepackage{latexsym}
\usepackage{amsmath}

\usepackage[T1]{fontenc}

\usepackage[utf8]{inputenc}

\usepackage{microtype}
\usepackage{graphicx}
\usepackage{booktabs}
\usepackage{array}
\usepackage{hyperref}       
\usepackage{url}            
\usepackage{booktabs}       
\usepackage{amsfonts}       
\usepackage{nicefrac}       

\usepackage{lipsum}		
\usepackage{graphicx}
\usepackage{natbib}
\usepackage{doi}
\usepackage{wrapfig}
\usepackage{subcaption} 
\usepackage{multirow} 

\usepackage{longtable}  
\usepackage{supertabular}
\usepackage{booktabs}  

\usepackage[utf8]{inputenc} 
\usepackage[T1]{fontenc}    
\usepackage{hyperref}       
\usepackage{url}            
\usepackage{amsfonts}       
\usepackage{nicefrac}       
\usepackage{microtype}      
\usepackage{doi}
\usepackage{xcolor}         
\usepackage{color}
\definecolor{mypurple}{RGB}{200, 125, 245}
\definecolor{myred}{RGB}{238, 34, 12}
\definecolor{mygrey}{RGB}{146, 146, 146}
\usepackage{tabularx}
\usepackage{makecell}
\usepackage{xspace}
\usepackage{enumitem}
\usepackage{float}
\usepackage{amsmath}
\usepackage{listings}
\usepackage{algorithm}
\usepackage{algorithmic}
\usepackage{mathrsfs}
\usepackage{longtable}
\usepackage{pythonhighlight}

\usepackage{supertabular,booktabs}
\usepackage[hang,flushmargin]{footmisc}

\title{Inductive-Deductive Strategy Reuse for Multi-Turn Instructional Dialogues}

\makeatletter
\def\thanks#1{\protected@xdef\@thanks{\@thanks
        \protect\footnotetext{#1}}}
\makeatother
\author{ 
Jiao Ou\textsuperscript{\rm 1*}\thanks{$^*$Jiao Ou and Jiayu Wu are co-first authors of this work.}, \textbf{Jiayu Wu\textsuperscript{\rm 1*}}, Che Liu\textsuperscript{\rm 1}, \\ \textbf{Fuzheng Zhang\textsuperscript{\rm 1}}, \textbf{Di Zhang\textsuperscript{\rm 1}}, \textbf{Kun Gai\textsuperscript{\rm 1}}\\ 
\textsuperscript{\rm 1} Kuaishou\\
{ojiao1111@gmail.com, jiayuwu9@163.com}
}

\begin{document}
\maketitle
\begin{abstract}
Aligning large language models (LLMs) with human expectations requires high-quality instructional dialogues, which usually require instructions that are diverse and in-depth.
Existing methods leverage two LLMs to interact for automatic collection: one simulating a user to pose instructions, and the other acting as a system agent to respond.
However, these user simulators struggle to model the rules behind how dialogues can pose different instructions without explicit guidance, resulting in general instructions.
In this paper, we propose to explicitly capture the complex rules to help the user simulator pose diverse and in-depth instruction.
Specifically, we first induce high-level instruction strategies from various real instruction dialogues serving as rules.
Afterward, different possible strategies are applied to the newly given dialogue scenario deductively to pose various instructions.
Experimental results show that our method can generate diverse and in-depth instructions. The constructed multi-turn instructional dialogues can outperform competitive baselines on the downstream chat model~\footnote{https://github.com/kwai/IDEAS}.
\end{abstract}

\section{Introduction}
Large language models (LLMs)~\cite{du2022glm,OpenAI2023GPT4TR} have demonstrated emergent capabilities across a wide range of language-related tasks naturally and interactively.
This mainly benefits from the alignment of LLMs with humans.
The alignment requires high-quality multi-turn instructions for fine-tuning LLMs in a dialogue-based setting~\cite{wang2023aligning}.
Such instructional dialogues usually require instructions that are diverse and in-depth~\cite{yu2016strategy,li2017learning}.
Specifically, diverse instructions help instruction-following LLMs adapt to different situations and user needs.
In-depth instructions help LLMs follow the logical flows of dialogues and deepen the understanding of dialogue histories to answer users' instructions.
However, manually collecting such instructional dialogue data is usually labor-intensive and time-consuming~\cite{wang-etal-2023-self-instruct}.
\begin{figure}
    \centering 
    \includegraphics[width=1\linewidth]{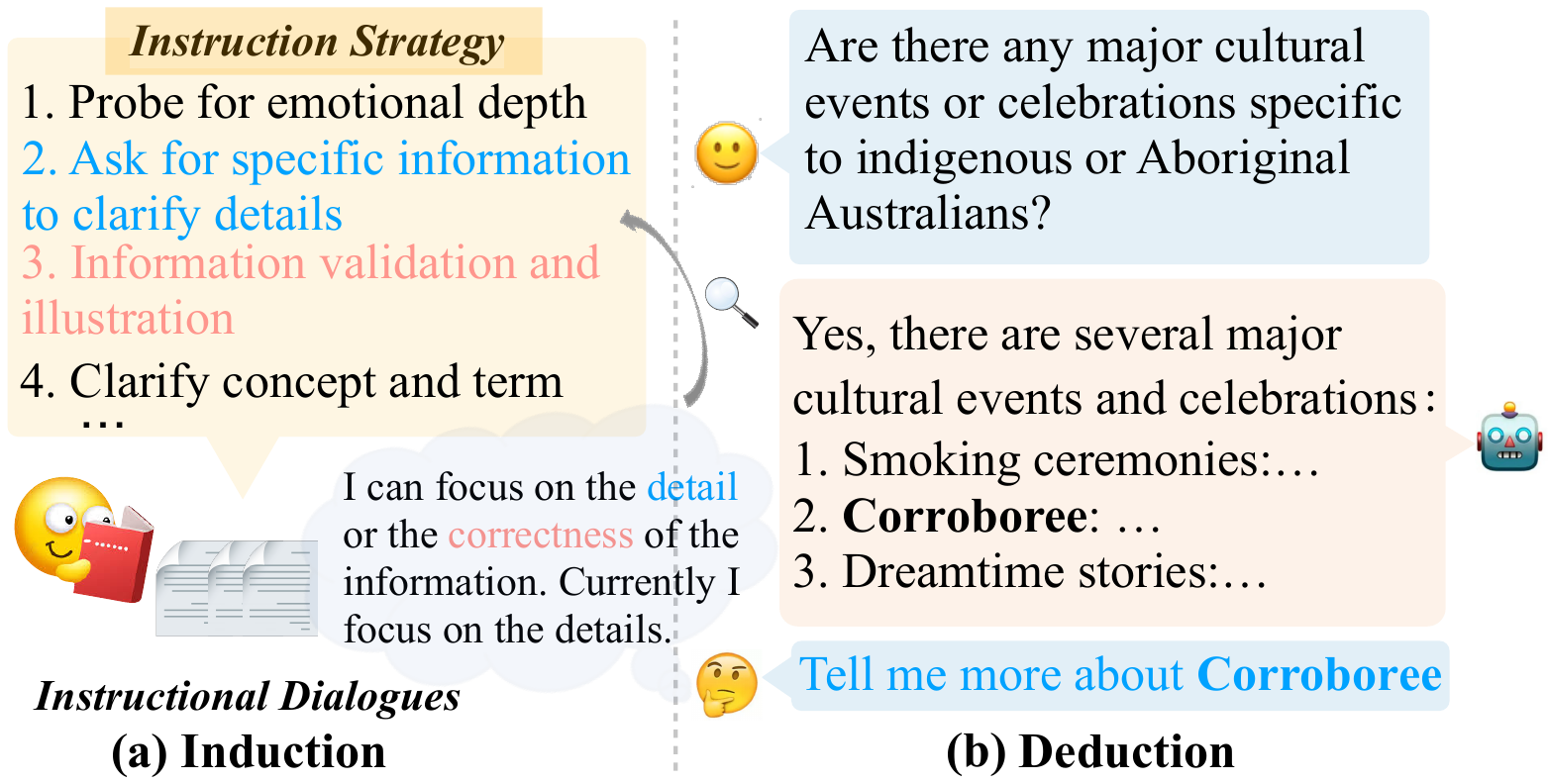}     
    \caption{An example of humans generating instructions by deductively utilizing instruction strategies, derived from inductive analysis of instructional dialogues.
    } 
    \label{fig:intro} 
\end{figure}

A feasible approach to automatic collection is leveraging two LLMs to interact: one simulating a user to pose instructions, and the other acting as a system agent to respond.
User simulators are currently implemented by employing role-played LLMs to simulate humans through instruction prompts~\cite{xu-etal-2023-baize,ding-etal-2023-enhancing} or fine-tuning LLMs to learn instruction generation from real instructional dialogues~\cite{kong2023platolm,sun2023parrot}.
However, dialogue data allows for different dialogue flows for the given dialogue history, i.e., various instructions can be posed~\cite{hou-etal-2018-sequence}.
LLMs in these user simulators implicitly learn the underlying rules guiding these different flows from training data.  
Due to the complexity and variability of these rules, LLMs struggle to capture various possible rules without explicit guidance so that the user simulators tend to capture frequently co-occurring patterns and generate more generic instructions~\cite{li2017learning}. 
This results in diminished diversity and insufficient depth of multi-turn instructions.

In this paper, we propose to explicitly model the rules behind how dialogues can flow in different directions, posing diverse and in-depth instructions.
Inspired by the cognitive abilities in human learning~\cite{lin1992self}, we observe that humans can gain general rules from various real dialogues via inductive reasoning ~\cite{hayes2010inductive}, which guides the dialogue flows in certain directions.
As shown in Figure~\ref{fig:intro}, humans induce rules such as ``Ask for specific information to clarify detail'' and ``Information validation and illustration''.
We refer to these rules as \emph{instruction strategies}.
Afterward, humans can pose various instructions in new dialogue scenarios by choosing different possible strategies via deductive reasoning~\cite{goel2007anatomy}.
Specifically, different strategies focus on different aspects, such as the ``detail'' or ``correctness'' of information, which directs the dialogue flows in various directions.

Motivated by this, we propose an \textbf{I}nductive-\textbf{De}ductive Str\textbf{a}tegy Reu\textbf{s}e method, IDEAS for short, to generate diverse and in-depth instructions for building multi-turn instructional dialogues.
IDEAS is composed of an induction stage and a deduction stage.
In the induction stage, we prompt GPT-4 to extract the instruction strategy for each pair of the dialogue history and the corresponding instruction in real human-machine dialogues.
Those similar instruction strategies are further abstracted into a high-level strategy, which reveals the generic nature regardless of the details relevant to the specific dialogues. 
In the deduction stage, the user simulator is asked to sample an appropriate instruction strategy to guide instruction generation based on the current dialogue history in each new dialogue scenario.
Afterward, the user simulator and a system agent interact iteratively to build multi-turn instructional dialogues.

Experiment results indicate that IDEAS can generate diverse and in-depth instructions, and our constructed multi-turn instructional dialogues contribute to the performance improvement of the downstream chat model. 
Our contributions are summarized as follows:
(1) 
To the best of our knowledge, this is the first study to propose an inductive-deductive strategy reuse method to generate diverse and in-depth instructions in new dialogue scenarios.
(2) Experimental results show that IDEAS produces higher-quality instructions, which can be further used to improve the performance of the downstream chat models.
(3) Extensive experiments show that providing more instructional dialogues with high-quality instructions can further improve performance.

\section{Preliminary}
This section describes task definitions and reviews the concepts of inductive and deductive reasoning.

\subsection{Task Definitions}
\label{appendix:definition}
In this paper, we first study the generation of high-quality instructions to construct multi-turn instructional dialogues and then use the constructed instruction dialogues to fine-tune the chat model to improve the model's ability to answer instructions.
Thus, this work involves two tasks, including \emph{instruction generation} and \emph{answer generation}.
\paragraph{Instruction Generation.}
We consider instruction generation tasks that require fine-tuning a model to generate reasonable instructions according to the given dialogue history.
More formally, we are given a set of training examples $\mathcal{D}_{ins}=\{(\boldsymbol{q}_0^i, \boldsymbol{a}_0^i, \dots, \boldsymbol{q}_T^i, \boldsymbol{a}_T^i)_{i=1}^N\}$, the generative model learns to model the distribution $\mathcal{P}_\phi(\boldsymbol{q}_t^i|\mathbf{h}_t^i)$ of the instruction $\boldsymbol{q}_t^i$ given the dialogue history $\mathbf{h}_t^i = \{\boldsymbol{q}_0^i, \boldsymbol{a}_0^i, \dots, \boldsymbol{a}_{t-1}^i\}$.
The model parameters $\phi$ can be learned by minimizing the following loss:
\begin{equation}
    \mathcal{L}_{i} = -\sum\nolimits_{i=1}^{N}\sum\nolimits_{t=1}^{T}\log P_\phi(\boldsymbol{q}_t^i|\mathbf{h}_t^i),
\end{equation}
This generative model can be a role-played LLM or a fine-tuned LLM on real human-machine dialogues as a user simulator, which iteratively interacts with a system agent (e.g., GPT-4) to obtain multi-turn instructional dialogues for different dialogue scenarios.

\paragraph{Answer Generation.}
We consider response generation tasks that require fine-tuning a chat model to generate reasonable responses according to the given dialogue history.
This chat model is fine-tuned on the generated multi-turn instructional dialogues for answering user's instructions.
Given the generated dialogues $\mathcal{D}_{chat}=\{(\boldsymbol{q}_0^i, \boldsymbol{a}_0^i, \dots, \boldsymbol{q}_T^i, \boldsymbol{a}_T^i)_{i=1}^M\}$, 
the parameter $\theta$ of the chat model can be optimized similarly as follows:
\begin{equation}
    \mathcal{L}_{c} = -\sum_{i=1}^{M}\sum_{t=1}^{T}\log P_\theta(\boldsymbol{a}_t^i|\boldsymbol{q}_0^i, \boldsymbol{a}_0^i, \dots, \boldsymbol{q}_t^i).
\end{equation}

\subsection{Induction \& Deduction}
\paragraph{Inductive Reasoning.}
Inductive reasoning, gaining common patterns and forming high-level rules from observations, is a core aspect of human intelligence~\cite{lake2017building}.
Formally, inductive reasoning is to infer an unknown rule $\boldsymbol{f}: \boldsymbol{\mathcal{X}} \to \boldsymbol{\mathcal{Y}}$ that maps an input $\boldsymbol{x} \in \boldsymbol{\mathcal{X}}$ to an output $\boldsymbol{y} \in \boldsymbol{\mathcal{Y}}$. The rule $\boldsymbol{f}$ can take various forms, such as natural language descriptions~\cite{wang2023hypothesis}.
For instruction generation, inductive reasoning is inferring an instruction strategy $\boldsymbol{f}$ 
from several pairs of the dialogue histories $\{\boldsymbol{q}_0, \boldsymbol{a}_0, \dots, \boldsymbol{a}_{t-1}\}$ and the corresponding instructions $\boldsymbol{q}_t$.
These dialogue histories often share common patterns and the rule $\boldsymbol{f}$ is prevalent in the mapping of all these pairs.

\paragraph{Deductive Reasoning.}
Deductive reasoning aims to derive new facts based on known facts and the induced rules~\cite{goel2007anatomy}.
Formally, deductive reasoning is to derive an output $\boldsymbol{y}'$ by applying $\boldsymbol{f}$ to a given input $\boldsymbol{x}'$, i.e., $\boldsymbol{y}'=\boldsymbol{f}(\boldsymbol{x}')$.
For instruction generation, deductive reasoning is deriving an instruction $\boldsymbol{q}'_t$ based on a new dialogue history $\mathbf{h}'_t =\{\boldsymbol{q}'_0, \boldsymbol{a}'_0, \dots, \boldsymbol{a}'_{t-1}\}$ and an instruction strategy, i.e., the induced rule  $\boldsymbol{f}$.

\begin{figure*}
\centering
\includegraphics[width=1\textwidth,keepaspectratio]{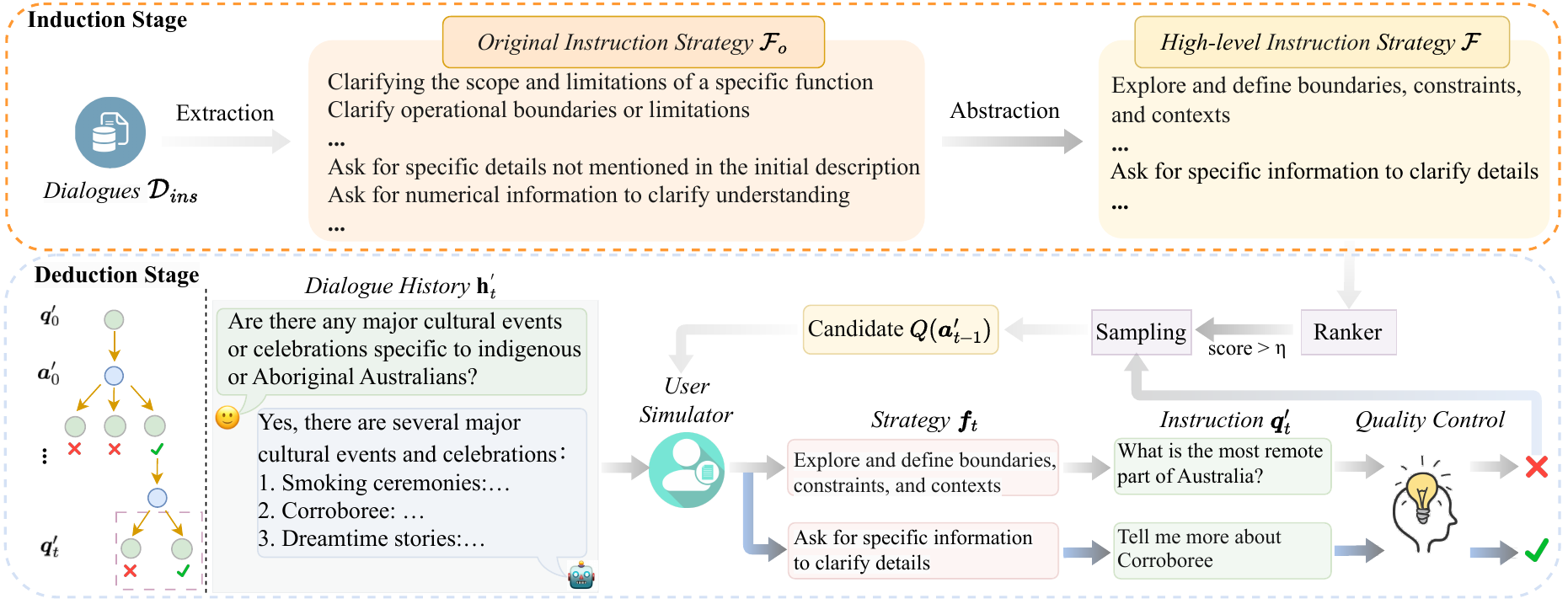}
\caption{The overall architecture of building multi-turn instructional dialogues.
In the induction stage, IDEAS induces high-level strategies $\mathcal{F}$ from human-machine instructional dialogues $\mathcal{D}_{ins}$.
In the deduction stage, the user simulator iteratively interacts with a system agent to produce new multi-turn dialogues based on the given opening line $\{\boldsymbol{q}'_0, \boldsymbol{a}'_0\}$, as shown on the left side. For generating the current instruction $\boldsymbol{q}'_t$ , the user simulator first chooses an appropriate strategy $\boldsymbol{f}_t$ from the candidate $Q(\boldsymbol{a}'_{t-1})$ based on the dialogue history $\mathbf{h}'_t$, and then generate $\boldsymbol{q}'_t$. If the quality does not meet the requirement, $\boldsymbol{q}'_t$ is regenerated.
This process is shown on the right side.
}
\label{fig:overall_architecture}
\end{figure*}
\section{IDEAS}
We aim to reuse instruction strategies to generate diverse and in-depth instructions for building multi-turn instructional dialogues. 
To this end, we introduce inductive reasoning for gaining high-level strategies from human-machine instructional dialogues $\mathcal{D}_{ins}$ in Section~\ref{app:induction}. 
In Section~\ref{app:deduction}, we describe how the user simulator chooses an appropriate strategy to generate instruction.
In Section~\ref{app:train}, we show the implementation of the involved models.
The overall architecture of building instructional dialogues is shown in Figure~\ref{fig:overall_architecture}.

\subsection{Induction Stage}
\label{app:induction}
This stage is to induce instruction strategies $\mathcal{F}$ from the given human-machine dialogues $\mathcal{D}_{ins}=\{(\boldsymbol{q}_0^i, \boldsymbol{a}_0^i, \dots, \boldsymbol{q}_T^i, \boldsymbol{a}_T^i)_{i=1}^N\}$.
Firstly, we successively extract each strategy expressed in natural language for each pair of dialogue history and the corresponding instruction.
However, there are several challenges in reusing:
(1) The strategies may involve details related to specific dialogues, rendering their reuse challenging~\cite{zheng2024stepback}.
(2) 
The variability and similarity of strategies further complicate the matter, as it is time-consuming to find the appropriate strategy for each dialogue history.
Some examples are shown in Figure~\ref{fig:overall_architecture}.
Therefore, we conduct high-level abstractions on these strategies to uncover their nature and reduce the strategy pool.
Next, we delve into instruction strategy extraction and abstraction in detail.

\paragraph{Instruction Strategy Extraction.}
For each pair of dialogue history $\mathbf{h}_t = \{\boldsymbol{q}_0, \boldsymbol{a}_0, \dots, \boldsymbol{a}_{t-1}\}$ and the corresponding instruction $\boldsymbol{q}_t$ from each dialogue in $\mathcal{D}_{ins}$, we prompt GPT-4 with a task description and the input-output pair to extract an instruction strategy $\boldsymbol{f}_o$ as follows:
\begin{equation}
    \boldsymbol{f}_o \sim P_{\text{Extraction}}(\cdot | \mathbf{h}_t, \boldsymbol{q}_t),
\end{equation}
where 
$\boldsymbol{f}_o$ is expressed in natural language. 
This prompt can be found in Table~\ref{tab:extraction_prompt}.
Consequently, we obtain original instruction strategies $\mathcal{F}_o$ on $\mathcal{D}_{ins}$.

\paragraph{Instruction Strategy Abstraction.}
We observe that many strategies bear similarities, they differ subtly in their expression or specific details.
We thus generalize those similar strategies into a high-level strategy.
A straightforward approach is to prompt GPT-4 to cluster original strategies $\mathcal{F}_o$ and then generalize for each cluster.
However, it is unrealistic to feed all strategies into GPT-4 due to the context length limitation~\cite{xu2023exploring}.
Besides, a single prompt cannot guide GPT-4 to handle complex tasks~\cite{li2023camel}.
Thus, we perform high-level abstraction in two steps, including strategy clustering and cluster generalization.

For strategy clustering, we cannot approximate the number of clusters in advance.
Thus, we take inspiration from~\citet{xu2023exploring} to form clusters by finding collections of similar strategies.
Each collection $\mathbf{C}_i$ contains the focused strategy $\boldsymbol{f}_{oi}$ and its similar strategies.
Specifically, we take each original strategy $\boldsymbol{f}_{oi}$ as the focused strategy in turn and retrieve similar strategies for each $\boldsymbol{f}_{oi}$ as follows:
\begin{equation}
\label{eq:epsilon}
\begin{aligned}
    \mathbf{C}_i = \{\boldsymbol{f}_{oi}\} \cup \{\boldsymbol{f}_{oj} | \operatorname{cos}(E(\boldsymbol{f}_{oi}), E(\boldsymbol{f}_{oj})) > \epsilon\}, \\
\end{aligned}
\end{equation}
where $\boldsymbol{f}_{oi}, \boldsymbol{f}_{oj} \in \mathcal{F}_o$ and $i \neq j$. 
$E(\cdot)$ denotes the embedding obtained by one Sentence-BERT model~\cite{song2020mpnet}~\footnote{Model name: all-mpnet-base-v2}, and $\operatorname{cos}(\cdot)$ denotes the cosine similarity.
$\epsilon$ represents the similarity threshold.
Note that $\mathbf{C}_k$ will not be obtained if the focused $\boldsymbol{f}_{ok}$ is already included in the formed collection $\mathbf{C}_i$.
In addition, we find that a certain $\boldsymbol{f}_{oj}$ may be similar to multiple strategies at the same time. 
To this end, we put $\boldsymbol{f}_{oj}$ into $\mathbf{C}_i$ corresponding to its most similar strategy $\boldsymbol{f}_{oj}$.
Consequently, we perform generalization for $K$ achieved clusters.

For cluster generalization,  we also prompt GPT-4 to generate a high-level instruction strategy for each cluster.
Please see Table~\ref{tab:abstraction_prompt} for this prompt.
The size of each collection $\mathbf{C}_i$ is considerably smaller than $\mathcal{F}_o$, and it is relatively simple to generalize a set of similar strategies.
Specifically, we prompt GPT-4 with task description and $\mathbf{C}_i$, and then output the high-level strategy $\boldsymbol{f}_i$ as follows:
\begin{equation}
    \boldsymbol{f}_i \sim P_{\text{Abstraction}}(\cdot | \mathbf{C}_i).
\end{equation}
Finally, the high-level instruction strategies $\mathcal{F} = \{\boldsymbol{f}_i\}_{i=1}^K$ are applied to new dialogue scenarios.

\subsection{Deduction Stage}
\label{app:deduction}
This stage is to guide the user simulator to generate diverse and in-depth instructions in a new dialogue scenario by reusing appropriate strategies in $\mathcal{F}$.
We initiate a new dialogue scenario by providing a new opening line, which contains an instruction-answer pair $\{\boldsymbol{q}'_0, \boldsymbol{a}'_0\}$.
Afterward, the user simulator iteratively interacts with a system agent to produce new multi-turn dialogues under the given opening line.
However, two sub-problems need to be addressed: (1) how to utilize the strategy; and (2) how to ensure the quality of the generated instruction.
Next, we describe the corresponding solutions.

\paragraph{Instruction Strategy Utilization.}
The natural idea is to leverage the chain-of-thought (CoT) technique~\cite{wei2022chain}, i.e., the user simulator first determines the appropriate instruction strategy based on the given dialogue history and then generates instructions.
However, it is as challenging as generating high-quality instructions directly since it also requires tracking and understanding the dialogue flows~\cite{ou-etal-2022-counterfactual}.
Inspired by~\citet{zhu2023large}, we provide a candidate set of instruction strategies, requiring the user simulator to retrieve a strategy from this set.
If all strategies $\mathcal{F}$ are used as the candidate, the user simulator also suffers from the context length limitation.
Thus, we roughly select $W$ potentially applicable strategies from $\mathcal{F}$ as the candidate.

Specifically, we introduce a ranker to calculate the probability that each instruction strategy $\boldsymbol{f}_i$ in $\mathcal{F}$ is suitable for the given dialogue history and consider those larger than a threshold $\eta$.
We further randomly sample $W$ strategies as the candidate.
More concretely, when generating the $t$-th instruction for the current dialogue history $\mathbf{h}'_t = \{\boldsymbol{q}'_0, \boldsymbol{a}'_0, \dots, \boldsymbol{a}'_{t-1}\}$  , the candidate $Q(\boldsymbol{a}'_{t-1})$ is built as follows:
\begin{equation}
\begin{aligned}
    Q'(\boldsymbol{a}'_{t-1}) &\leftarrow \{\boldsymbol{f}_i | \operatorname{Ranker}(\boldsymbol{a}'_{t-1}, \boldsymbol{f}_i) > \eta\}, \\
    Q(\boldsymbol{a}'_{t-1}) &\leftarrow \operatorname{Sample}(Q'(\boldsymbol{a}'_{(t-1)}), W), \\
\end{aligned}
\label{eq:candidate}
\end{equation}
where $\boldsymbol{f}_i \in \mathcal{F}$ and $\boldsymbol{a}'_{t-1}$ is the $(t-1)$-th answer (more details about the usage of $\boldsymbol{a}'_{t-1}$ rather than $\mathbf{h}'_t$ in Section~\ref{app:ranker}).
Further, we provide the task description, the dialogue history $\mathbf{h}'_t$, and the candidate $Q(\boldsymbol{a}'_{t-1})$ as the input.
The input is used to prompt the user simulator to sample the selected strategy $\boldsymbol{f}_t$ and the instruction $\boldsymbol{q}'_t$ as successively
\begin{equation}
   \boldsymbol{u}_t  \sim P_\phi(\cdot | \mathbf{h}'_t, Q(\boldsymbol{a}'_{t-1})),
\label{eq:generate}
\end{equation}
where $\boldsymbol{u}_t$ is a concatenated string with the form of [instruction strategy]$\boldsymbol{f}_t$[instruction]$\boldsymbol{q}'_t$.
This prompt can be found in Table~\ref{tab:user_prompt}.

\paragraph{Quality Control.}
Low-quality instructions are inevitable when generating instructions.
We design a \emph{reflection} module to judge the quality of the generated instruction immediately.
If the quality meets the requirement, the interaction continues, i.e., the system agent generates the answer $\boldsymbol{a}'_t$ to $\boldsymbol{q}'_t$; otherwise, the instruction $\boldsymbol{q}'_t$ is regenerated.
Considering that the low quality of the instruction may be related to the selected strategy, the candidate $Q(\boldsymbol{a}'_{t-1})$ will be randomly resampled and no longer include the previously selected strategies when regenerating.
Therefore, the core issue is how to judge whether the quality meets the requirement.

Specifically, we refer to widely-used metrics for dialogue evaluation~\cite{mehri-eskenazi-2020-usr}, along with our goal, i.e., generating diverse and in-depth instructions. We finally choose two appropriate metrics, \emph{correctness} and \emph{coherence}, to evaluate instructions.
Correctness requires that the generated instruction $\boldsymbol{q}'_t$ cannot contradict the dialogue history $\mathbf{h}'_t$ and cannot be answered by existing answers $\{\boldsymbol{a}'_0, \dots, \boldsymbol{a}'_{t-1}\}$, which can avoid talking about highly similar topics and promote the progress of the dialogue.
Accordingly, coherence requires that $\boldsymbol{q}'_t$ is related to $\mathbf{h}'_t$, and coherently connected to the previous instructions or answers.
The coherent instruction can lead to in-depth interaction.
Since there are no automatic evaluation metrics available, we still prompt GPT-4 to judge instructions from these two dimensions.
The corresponding prompt can be found in Table~\ref{tab:qc_prompt}.
Finally, we obtain the constructed multi-turn dialogues $\mathcal{D}_{chat}$ for training the chat model.
Please refer to~\ref{appendix:detail} for the corresponding training details.

\subsection{Model Implementation}
\label{app:train}

\paragraph{User Simulator.}
We follow~\citet{kong2023platolm} and~\citet{sun2023parrot} to choose the fine-tuned open-source LLM as the user simulator due to its lightweight and free.
Specifically, we fine-tune LLaMA-2~\cite{touvron2023llama} on human-machine instructional dialogues $\mathcal{D}_{ins}$ to learn $P_{\phi}(\boldsymbol{U} | \mathbf{H}, \boldsymbol{Q})$.
The input is a prompt in Table~\ref{tab:user_prompt} with the task description, the dialogue history $\mathbf{H}$, and the corresponding candidate $\boldsymbol{Q}$.
The output is the selected instruction strategy and the instruction $\boldsymbol{U}$.
We minimize the objective as follows:
\begin{equation}
    \mathcal{L}_{user} = -\sum\nolimits_{z=1}^{|U|}\log P_\phi(U_z|\mathbf{H}, \boldsymbol{Q}, \boldsymbol{U}_{<z}),
\end{equation}
where $\boldsymbol{U}_{<z}$ is a prefix of the concatenated string with the strategy and instruction. $|U|$ denotes the length of $\boldsymbol{U}$.
$\boldsymbol{Q}$ contains the golden strategy and $(W-1)$ randomly sampled strategies from $\mathcal{F}$.

\paragraph{Ranker.}
\label{app:ranker}
We build a classification dataset $\mathcal{D}_{r} = \{(\mathbf{h}^i, \boldsymbol{f}^i, l^i)_{i=1}^R\}$ based on $\mathcal{D}_{ins}$ and the corresponding $\mathcal{F}$, where $\mathbf{h}^i$ is a dialogue history and $\boldsymbol{f}^i$ is an instruction strategy. 
$l^i \in \{0, 1\}$ denotes a matching label, which indicates that $\boldsymbol{f}^i$ is an appropriate strategy for $\mathbf{h}^i$ if $l^i = 1$, otherwise $l^i = 0$.
For simplicity, the superscript $i$ will be omitted in the following.
We fine-tune BERT~\cite{devlin-etal-2019-bert} on $\mathcal{D}_r$ to learn $P_{\psi}(l|\mathbf{H}, \boldsymbol{F})$ for correctly identifying the positive strategy from a set of negative strategies.
We minimize the following objective,
\begin{equation}
\begin{aligned}
    \mathcal{L}_{r} = & -[l\log P_\theta(l=1|\mathbf{H}, \boldsymbol{F}) \\
    & + (1-l)\log P_\theta(l=0|\mathbf{H}, \boldsymbol{F})].
\end{aligned}
\end{equation}
For a given dialogue history $\mathbf{h}^i$ in $\mathcal{D}_{ins}$, the positive is the high-level strategy $\boldsymbol{f}^i$ obtained from the subsequent instruction of $\mathbf{h}^i$ in $\mathcal{D}_{ins}$, while the negative is randomly selected from $\mathcal{F}$.
One dialogue history has one positive strategy and one negative strategy.
Due to the BERT's context length limitation, we only select the last-turn answer to represent the entire dialogue history.
This ranker may have limited performance, but this lightweight model is sufficient for a coarse selection of candidates.
Please see more analysis about this design in Appendix~\ref{appendix:ranker}.

\section{Experimental Setup}
\subsection{Settings}
The basic data used for all experiments are collected by~\citet{sun2023parrot}.
For instruction generation, we use the human-machine instructional dialogues from ShareGPT to obtain instruction strategies, along with training the user simulator and the ranker in the induction stage. 
In the deduction stage, we randomly sample 10K dialogues from unused ShareGPT and UltraChat~\cite{ding-etal-2023-enhancing} datasets, and then intercept the first-turn $\{\boldsymbol{q}'_0, \boldsymbol{a}’_0\}$ as opening lines.
Please see Appendix~\ref{appendix:details} for more details on data and method implementations.

\subsection{Baselines}
\label{setup:baseline}
We compare IDEAS with a set of baselines:
(1) \textbf{Self-Chat}~\cite{xu-etal-2023-baize}, which uses the instructions from opening lines as seeds and prompts GPT-4 to generate transcripts for both sides of the dialogues until a natural stopping point is reached.
(2) \textbf{Iterative Self-Chat (Iter Self-Chat)}~\cite{ding-etal-2023-enhancing}, which adopts two separate GPT-4 to generate dialogues based on the given opening lines. One simulates the use in generating instructions, and the other generates the answer.
(3) \textbf{Parrot-Ask}~\cite{sun2023parrot}, which fine-tunes LLaMA-2 as the user simulator. It takes the dialogue history as the input to generate instructions.
(4) \textbf{SkillGen}, which is a variant of IDEAS. The input is the given dialogue history and the output is the sequentially generated instruction strategy and instruction.
The difference is that IDEAS provides strategy candidates for the user simulator to select rather than generate the appropriate strategy.
Note that the system agents in all baselines adopt GPT-4.
Those baselines using fine-tuned LLaMA-2 as user simulators also introduce our reflection module for quality control.
For a fair comparison, all experimental setups of baselines are consistent with IDEAS.
We also test three different abstraction level in Eq.~\ref{eq:epsilon}, i.e., $\epsilon = 0.4, 0.5, 0.6$.
In addition, we supplement \textbf{GPT-4} and \textbf{ChatGPT} as chat models to clarify the room for improvement.

\begin{table*}[t!]
\centering
\small
\begin{tabular}{lcccccccccccc}
\toprule
          \multicolumn{1}{l}{\multirow{2}{*}{\textbf{Method}}} & \multicolumn{5}{c}{\textbf{ShareGPT}}                                                                                                                 & \multicolumn{5}{c}{\textbf{UltraChat}}                                                                                                                                    \\
           & \textbf{Appr.} & \multicolumn{1}{c}{\textbf{Coh.}} & \multicolumn{1}{c}{\textbf{Dep.}} & \multicolumn{1}{c}{\textbf{Ins.}} & \multicolumn{1}{c}{\textbf{Div.}} & \multicolumn{1}{c}{\textbf{Appr.}} & \multicolumn{1}{c}{\textbf{Coh.}} & \multicolumn{1}{c}{\textbf{Dep.}} & \multicolumn{1}{c}{\textbf{Ins.}} & \multicolumn{1}{c}{\textbf{Div.}} \\ \toprule
Self-Chat   & \textbf{9.94}$^{\ \,}$             & \textbf{9.91}$^{\ \,}$                           & 2.65$^\ddagger$                         & 2.49$^\ddagger$                        & 2.11$^\ddagger$                          & \textbf{9.96}$^{\ \,}$                                & \textbf{9.81}$^{\ \,}$                         & 2.77$^\ddagger$                         & 2.59$^\ddagger$                        & 2.28$^\ddagger$                          \\
Iterative Self-Chat  & 9.65$^{\ \,}$            & 9.66$^{\ \,}$                         & 3.18$^\ddagger$                         & 3.04$^\ddagger$                        & 3.05$^\ddagger$                          & 9.61$^{\ \,}$                                & 9.53$^{\ \,}$                         & 3.42$^\ddagger$                         & 3.19$^\ddagger$                        & 3.17$^\ddagger$                          \\
Parrot-Ask & 7.89$^\ddagger$            & 7.71$^\ddagger$                          & 5.19$^\ddagger$                         & 4.62$^\ddagger$                        & 4.46$^\ddagger$                          & 7.74$^\ddagger$                                & 7.47$^\ddagger$                          & 5.89$^\ddagger$                         & 5.29$^\ddagger$                        & 5.16$^\ddagger$                          \\
SkillGen   & 8.05$^\ddagger$            & 7.91$^\ddagger$                          & 5.41$^\ddagger$                         & 4.88$^\ddagger$                        & 4.61$^\ddagger$                          & 8.40$^\ddagger$                                 & 8.23$^\ddagger$                          & 6.46$^\ddagger$                         & 5.77$^\ddagger$                        & 5.61$^\ddagger$                          \\ \midrule
\textbf{IDEAS} $\epsilon=0.6$    & 9.12$^\ddagger$              & 8.91$^\ddagger$                            & 6.57$^\ddagger$                           & 5.99$^\ddagger$                          & 5.51$^\ddagger$                            & 9.30$^\ddagger$                                 & 9.08$^\ddagger$                           & 7.41$^\ddagger$                           & 6.77$^\ddagger$                         & 6.68$^\ddagger$                            \\
\textbf{IDEAS} $\epsilon=0.5$   & 9.48$^{\ \,}$            & 9.26$^{\ \,}$                          & \textbf{6.89}$^{\ \,}$                         & \textbf{6.27}$^{\ \,}$                        & \textbf{5.85}$^{\ \,}$                          & 9.57$^{\ \,}$                                & 9.37$^{\ \,}$                          & \textbf{7.74}$^{\ \,}$                         & \textbf{7.11}$^{\ \,}$                        & \textbf{6.92}$^{\ \,}$                          \\ 
\textbf{IDEAS} $\epsilon=0.4$     & 9.07$^\ddagger$              & 8.86$^\ddagger$                            & 6.59$^\ddagger$                           & 5.92$^\ddagger$                          & 5.59$^\ddagger$                            & 9.26$^\ddagger$                                  & 8.96$^\ddagger$                            & 7.38$^\ddagger$                           & 6.63$^\ddagger$                         & 6.56$^\ddagger$                           \\ \midrule
$\mathcal{D}_{ins}$ (human)   & 8.87$^{\ \,}$            & 8.78$^{\ \,}$                          & 6.21$^{\ \,}$                         & 5.76$^{\ \,}$                        & 5.84$^{\ \,}$                          & -                                   & -                             & -                            & -                           & -      \\ \bottomrule                      
\end{tabular}
\caption{Automatic evaluation of the generated instructions by different methods on ShareGPT and UltraChat dialogue scenarios.
    The bottom row corresponds to the human-machine instructional dialogues in $\mathcal{D}_{ins}$.
    Significance tests between IDEAS ($\epsilon=0.5$) and baselines are performed using t-test.
     $\ddagger$ indicates $p$-value < 0.01. 
    }
\label{tab:data}
\end{table*}

\subsection{Evaluation Metrics}
\paragraph{Instruction Evaluation.}
Efficient instruction evaluation about depth and diversity is significantly understudied.
Thus, we follow~\citet{ding-etal-2023-enhancing} to design five metrics to evaluate generated instructions and implement each based on ``LLMs as Judges''~\cite{zheng2024judging} by prompting the top-tier LLM, i.e., \texttt{gpt-4-1106-preview} to output scores~\footnote{GPT-4 prefers to generate instructions for itself as the system agent, which may lead to inflated scores for Self-Chat and iter Self-Chat. However, it does not affect the ranking of these methods in generating diverse and in-depth instructions from the experimental results in Section~\ref{exp:instruction} since Self-Chat and Iter Self-Chat achieve the lowest scores on related metrics.}.
The prompt is shown in Table~\ref{tab:data_eval_prompt}.
(1) \textbf{Appropriateness (Appr.)}, which means whether the instruction is appropriate with the given dialogue history and should not be detached from the history.
(2) \textbf{Coherence (Coh.)}, which means whether the instruction is logically coherent to the given dialogue history.
(3) \textbf{Depth (Dep.)}, which means whether the instruction expands the topic or explores more details in the dialogue history.
(4) \textbf{Insight (Ins.)}, which means whether the instruction can bring new understanding, stimulate thought, lead to deeper interaction, or help unearth more valuable information, instead of repeating known content. 
(5) \textbf{Diversity (Div.)}, which means whether the instruction is different from the previous instructions. The difference is reflected in the instruction type.
The rating scale is of $1$ to $10$, in which $1$ means worst and $10$ best.

\paragraph{Model Evaluation.}
We use the following widely-used benchmarks to evaluate chat models.
(1) \textbf{AlpacaEval (AE)}~\cite{dubois2023alpacafarm}, which is an automatic evaluator for the single-turn instruction-following ability.
(2) \textbf{MT-Bench (MB)}~\cite{zheng2023judging}, which contains two-turn instructions that evaluate a chatbot’s multi-turn conversational and instruction-following ability;
(3) \textbf{MT-Bench++~ (MB++)}~\cite{sun2023parrot}, which expands the dialogues in MT-Bench to create an eight-turn evaluation dataset.
(4) \textbf{MT-Eval (ME)}~\cite{kwan2024mteval}, which is a comprehensive benchmark to evaluate multi-turn dialogue abilities.
The statistics of these benchmarks are shown in Table~\ref{tab:statistics}.

\section{Results and Discussion}
\subsection{Instruction Evaluation}
\label{exp:instruction}
We first evaluate the generated instructions.
To investigate the consistency between GPT-4 Judger and human annotators, we randomly select $500$ instructions for human evaluation.
Specifically, We employ three annotators to rate the instructions and present the same evaluation instruction as GPT-4.
The consistency is measured via the Fleiss's kappa $\kappa$~\cite{randolph2005free}. 
The $\kappa$ values for \emph{Appropriateness}, \emph{Coherence}, \emph{Depth}, \emph{Insight}, \emph{Diversity} are 0.71, 0.67, 0.59, 0.53, and 0.64 respectively.

The results are shown in Table~\ref{tab:data}, indicating that our generated instructions outperform all the baselines on the key optimized metrics.
We further observe that:
(1) Our generated instructions achieve scores similar to $\mathcal{D}_{ins}$ on all metrics, indicating that our generated instructions are high-quality.
We present the constructed dialogues in Table~\ref{tab:parrot_case} and~\ref{tab:ideas_case}.
(2) IDEAS achieves higher scores of \emph{Dep.}, \emph{Ins.} and \emph{Div.}, indicating that IDEAS can generate more diverse and in-depth instructions.
In particular, Parrot-Ask vs. IDEAS shows the effectiveness of inductive-deductive strategy reuse.
(3) SkillGen vs. IDEAS indicates the effectiveness of the provided candidate in strategy utilization.
(4) Self-Chat achieves the highest scores on \emph{Appr.} and \emph{Coh.}. We speculate that GPT-4 manages the generation of the entire dialogue, which may perform better in generating appropriate and coherent instructions. 
Besides, the scores are close to the full score. This is also partly because GPT-4 tends to give higher scores than the actual scores when scoring the text that it generates. 
Overall, IDEAS achieves relatively good scores of \emph{Appr.} and \emph{Coh.}
(5) The threshold $\epsilon=0.5$ achieves higher scores on all the metrics. Please refer to Section~\ref{app:abstract} for further analysis. Thus, $0.5$ can be used for building high-quality instructional dialogues.

\begin{table}[t!]
\centering
\small
\begin{tabular}{lcc}
\toprule
               & \textbf{Avg.\#Turns} & \textbf{Avg.\#Tokens}   \\
               \toprule
Self-Chat      & 3.10         & 28.53                                \\
Iterative Self-Chat & 8.08         & 31.38                                 \\
Parrot-Ask     & 9.32         & 36.85                                 \\
SkillGen       & 9.37         & 36.29                                 \\
IDEAS          & 9.33         & 36.90                                 \\
\bottomrule
\end{tabular}
\caption{Data statistics. Avg.\#Turns and Avg.\#Tokens denote the average turns of constructed instructional dialogues and the average length of generated instructions.
    }
\label{tab:statistic}
\end{table}

\begin{table}[t!]
\centering
\small
\begin{tabular}{llcccc}
\toprule
                            & \textbf{Method}                       & \textbf{AE}                  & \textbf{MB} & \textbf{MB++} & \textbf{ME} \\ \toprule 
                            & GPT-4                         & 95.28$^{\ \,}$                        & 8.99$^{\ \,}$                         & 9.18$^{\ \,}$                           & 9.03$^{\ \,}$                        \\
                            & ChatGPT                      & 89.37$^{\ \,}$                        & 7.94$^{\ \,}$                         & 8.33$^{\ \,}$                           & 7.72$^{\ \,}$                        \\ \midrule
                            & Self-Chat                   & 58.39$^\ddagger$                         & 6.15$^\ddagger$                          & 5.89$^\ddagger$                            & 7.02$^\ddagger$                         \\
                            & Iter Self-Chat                    & \textbf{87.59}$^{\ \,}$ & 6.78$^\dagger$                          & 6.83$^\ddagger$                           & 7.01$^\ddagger$                         \\
                            & Parrot-Ask                   & 77.89$^\dagger$                         & 6.71$^\ddagger$                          & 6.78$^\ddagger$                            & 7.07$^\ddagger$                         \\
                            & SkillGen                     & 78.26$^\dagger$                         & 6.79$^\dagger$                          & 6.85$^\dagger$                            & 7.17$^\dagger$                         \\
\multirow{-5}{*}{(a)}  & \textbf{IDEAS} & 78.39$^{\ \,}$                        & \textbf{6.92}$^{\ \,}$                         & \textbf{7.02}$^{\ \,}$                           & \textbf{7.25}$^{\ \,}$                        \\ \midrule
                            & Self-Chat                     & 44.58$^\ddagger$                         & 6.06$^\ddagger$                          & 6.28$^\ddagger$                            & 6.43$^\ddagger$                         \\
                            & Iter Self-Chat                    & \textbf{85.90}$^{\ \,}$                        & 6.30$^\ddagger$                         &  6.67$^\dagger$   & 6.40$^\ddagger$                        \\
                            & Parrot-Ask                   & 74.66$^\ddagger$                        & 6.38$^\dagger$                         & 6.57$^\ddagger$                           & 6.54$^\ddagger$                        \\
                            & SkillGen                     & 76.84$^\dagger$                        & 6.54$^\dagger$                         & 6.71$^\dagger$                           & 6.62$^\ddagger$                        \\
\multirow{-5}{*}{(b)} & \textbf{IDEAS} &  76.99$^{\ \,}$ & \textbf{6.63}$^{\ \,}$  &  \textbf{6.81}$^{\ \,}$   & \textbf{6.76}$^{\ \,}$ \\ \bottomrule 
\end{tabular}
\caption{Automatic evaluation of chat models trained on different constructed instructional dialogues on (a) ShareGPT and (b) UltraChat dialogue scenarios. $\dagger$ and $\ddagger$ indicate $p$-value < 0.05 and 0.01 respectively (significance tests via t-test). 
}
\label{tab:model}
\end{table}

\subsection{Model Evaluation}
We further evaluate the benefit of the constructed dialogues on the downstream chat models.
Table~\ref{tab:statistic} first shows data statistics.
The results are shown in Table~\ref{tab:model}, which indicates that IDEAS outperforms all the baselines on almost all the benchmarks.
This confirms the effectiveness of generating diverse and in-depth instructions.
We further observe that:
(1) IDEAS achieves higher scores for all multi-turn benchmarks compared to baselines, especially Parrot-Ask. 
This demonstrates that strategy reuse is effective for improving multi-turn dialogue and instruction-following abilities.
(2) Iter Self-Chat achieves the highest score on AlpacaEval which focuses on the single-turn instruction-following ability, but performs relatively poorly on the multi-turn benchmarks. 
This indicates that the instructions generated by the role-played GPT-4 in multi-turn interaction are weakly correlated, i.e., insufficient depth of multi-turn instructions.

\subsection{Further Discussion}
We investigate the impact of the abstraction level, the effect of each component of IDEAS, and the impact of the amount of instructional dialogues.

\paragraph{The Impact of Abstraction Level.}
\label{app:abstract}
We set the similarity threshold $\epsilon$ to different values, i.e., $0.4$, $0.5$, $0.6$, for assessing the impact of the abstraction level of instruction strategies. 
The results are shown in Table~\ref{tab:data}. 
We can observe that the scores of the generated instructions with the threshold $0.5$ reach a peak on all metrics, and drop as the level of abstraction increases (i.e., $0.4$).
This suggests that moderate abstraction that removes unnecessary details and distills high-level strategies is optimal.
We speculate that insufficient abstraction leads to strategies that involve details related to specific dialogues, i.e., are overly specific. 
However, these details may lead to strategies that do not fully match a new dialogue history.
Thus, these seemingly appropriate strategies are difficult to effectively utilize by user simulators, which tend to generate more general instructions.
Furthermore, excessive abstraction makes one strategy applicable to more dialogue histories, i.e., one strategy can guide multiple dialogue flows.
This still results in diminished diversity and insufficient depth.

\begin{table}[t!]
\centering
\small
\begin{tabular}{@{}lccc@{}}
\toprule

\textbf{Method} & \textbf{MT-Bench} & \textbf{MT-Bench++} & \textbf{MT-Eval}  \\
\midrule
IDEAS &  6.92$^{\ \,}$ & 7.02$^{\ \,}$ & 7.25$^{\ \,}$ \\ \midrule
w/o Reflection & 6.68$^\ddagger$  & 6.84$^\dagger$  & 7.21$^{\ \,}$  \\
w/o CandGen & 6.79$^\dagger$  & 6.85$^\dagger$  & 7.17$^\dagger$   \\
w/o Ranker & 6.81$^\dagger$  & 6.87$^\dagger$  & 7.16$^\dagger$  \\
w/o CandTop1 & 6.89$^{\ \,}$ & 6.92$^\dagger$  & 7.04$^\ddagger$   \\
w/o CandRand & 6.78$^\dagger$  & 6.97$^{\ \,}$ & 7.17$^\dagger$   \\
w/o Abstraction & 6.73$^\ddagger$  & 6.81$^\ddagger$  & 7.13$^\ddagger$   \\
\bottomrule
\end{tabular}
\caption{Ablation study of different components of IDEAS on the downstream chat model under the ShareGPT dialogue scenario. 
$\dagger$ and $\ddagger$ indicate $p$-value < 0.05 and 0.01 respectively (significance tests via t-test). 
}
\label{tab:ablation}

\end{table}
\paragraph{Ablation Study.}
We perform the following ablation tests to validate the effect of each component:
(1) Do not guarantee the quality of each instruction via quality control (w/o Reflection).
(2) Directly generate a strategy without the candidate for selection (w/o CandGen).
(3) Randomly choose $W$ strategies from $\mathcal{F}$ as the candidate without ranking (w/o Ranker).
(4) Provide the top-1 strategy via ranker to the user simulator without secondary selection (w/o CandTop1).
(5) Provide a strategy that randomly selects from the candidate without secondary selection (w/o CandRand).
(6) Select $W$ original strategies from $\mathcal{F}_o$ to replace the high-level strategies as the candidate (w/o Abstraction).
The results are shown in Table~\ref{tab:ablation}. 
We observe that ablating each component brings varying degrees of performance drop.
This demonstrates the necessity of designing all these components.

\begin{figure}
    \centering 
    \includegraphics[width=1\linewidth]{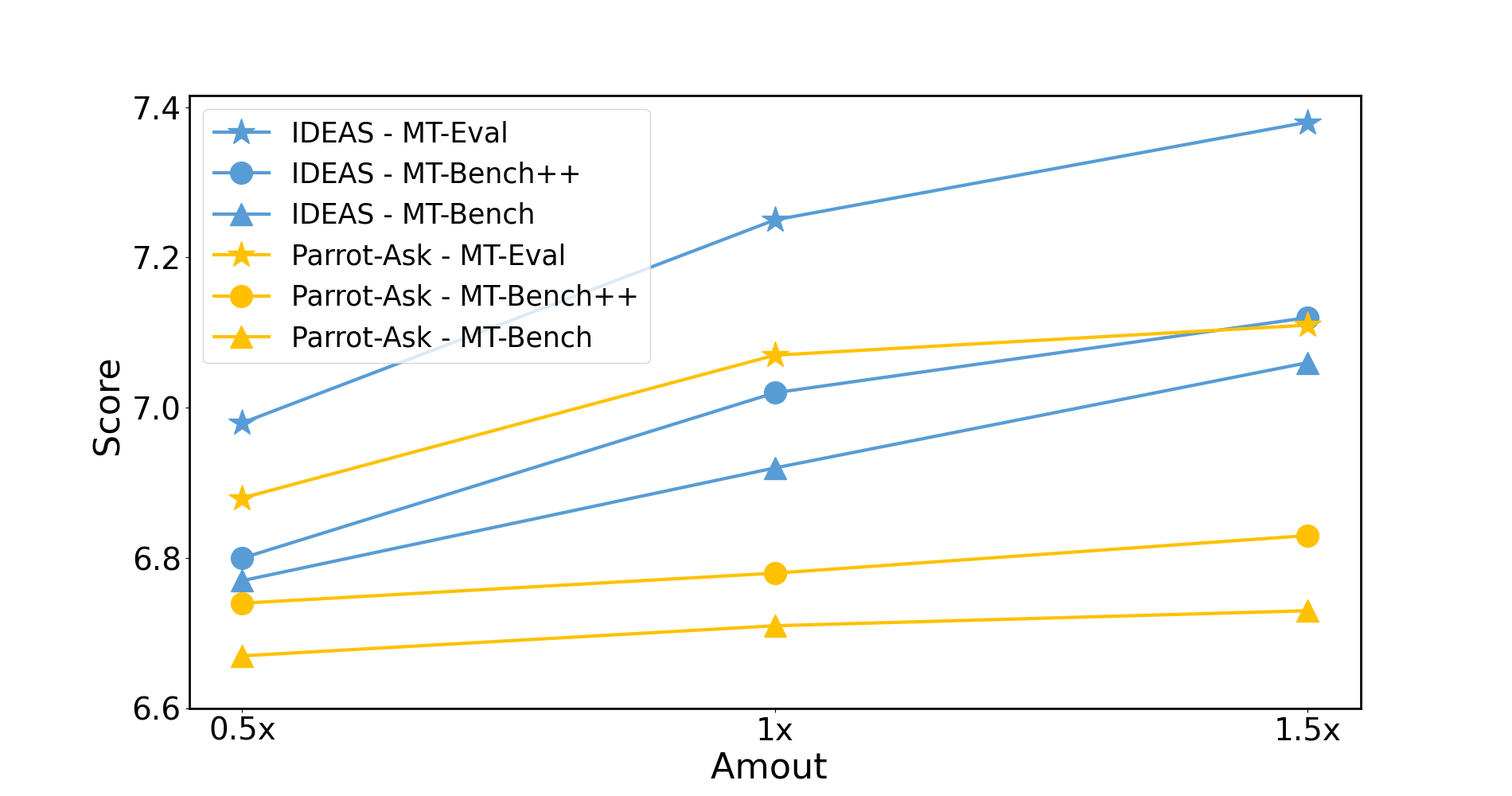}     
    \caption{Performance changes on chat models respectively by providing different amounts of instructional dialogues generated by IDEAS and Parrot-Ask. 
    } 
    \label{fig:amount} 
\end{figure}

\paragraph{The Impact of Amount.}
We select $0.5$x, $1$x, and $1.5$x the amount of generated dialogues to assess the impact of providing more multi-turn instructional dialogues and compare IDEAS with Parrot-Ask.
Note that $1$x represents that 1*10K generated multi-turn instructional dialogues are selected.
Therefore, we select an additional 5K opening lines from unused ShareGPT data~\cite{sun2023parrot} and then apply IDEAS and Parrot-Ask to generate more dialogues.
The results are shown in Figure~\ref{fig:amount}.
We can observe that:
(1) 
The scores of Parrot-Ask on all benchmarks reach a peak at 1x and no increase afterward.
We speculate that the Parrot-Ask struggles to implicitly capture complex rules, which leads to more generic instructions.
Further, similar dialogues at high amounts would not positively affect training~\cite{ou-etal-2022-counterfactual}.
(2) In contrast, the scores of IDEAS keep increasing from 0.5x to 1.5x.
This indicates that the multi-turn instructional dialogue data constructed by IDEAS is more diverse, further improving the performance of the downstream chat model.

\section{Related Work}
\paragraph{Instructional Dialogue Construction.}
Aligning LLMs with human expectations requires massive high-quality instructions.
With the success of top-tier LLMs, it is feasible to use strong LLMs to automatically construct a variety of instructions, e.g., single-turn instructions~\cite{wang-etal-2023-self-instruct,lamini-lm,gpt4all,honovich-etal-2023-unnatural,yu2023large,xu2023wizardlm} and multi-turn instructions~\cite{ding-etal-2023-enhancing,xu-etal-2023-baize,li2023camel,ji2023towards}.
The latest framework for constructing multi-turn instructions is leveraging two LLMs to interact: one simulating a user to pose instructions, and the other acting as a system agent to respond.
Our proposed method is used to improve the performance of this framework of multi-turn instructional dialogue construction.

\paragraph{Instructions Generation.}
The user simulator currently uses two implementations to generate instructions. One is to employ role-played LLMs to simulate humans through instruction prompts~\cite{xu-etal-2023-baize,ding-etal-2023-enhancing}. The other is to fine-tune LLMs to learn instruction generation from real instructional dialogues~\cite{kong2023large,sun2023parrot}.
Besides, some works in similar tasks study how to ask high-quality questions, including question generation in open-domain dialogue systems~\cite{wang2018learning,shen2021gtm} and conversational question generation~\cite{gu-etal-2021-chaincqg,li2022consecutive,zeng2023synthesize}.
These works capture the underlying rules that guide the dialogue flows in different directions.
In contrast, IDEAS explicitly models these underlying rules by introducing instruction strategies, to guide the dialogue flow in different directions.

\paragraph{Dialogue Flow Modeling.}
Some researches~\cite{ou-etal-2022-counterfactual,bao-etal-2023-synthetic} utilize a sequence of knowledge pieces (e.g., keywords, Wikipedia pages) to simulate the trajectory of dialogue flows for guiding open-domain dialogue generation.
Compared to existing works, we are the first to leverage abstract rules, i.e., high-level instruction strategies, for dialogue flow guidance.
We posit that LLMs have the capability to identify keywords and related information within extensive corpora. However, LLMs encounter difficulty in deriving underlying general rules. 
Thus, introducing high-level instruction strategies can potentially serve as a more effective direction, supporting the model in generating superior multi-turn dialogues.

\paragraph{Inductive Learning.}
Our work is based on inductive learning~\cite{michalski1983theory}, which has shown promising results in various scenarios, including reasoning tasks~\cite{wang2023hypothesis,zhu2023large,sun2024itd} and autonomous agents~\cite{park2023generative,zhao2023expel}.
However, our work focuses on inducing high-level instruction strategies from human-machine dialogues and then applying them to guide instruction generation in new dialogue scenarios, which has not been explored in instructional dialogue construction.

\section{Conclusion}
This paper presents an inductive-deductive strategy reuse method, IDEAS, to generate diverse, in-depth instructions for building multi-turn instructional dialogues.
Specifically, IDEAS first induces high-level instruction strategies from real dialogues via inductive reasoning.
Afterward, IDEAS deductively applies these strategies to new dialogue scenarios, where the strategies serve as underlying rules to guide the dialogue flow in different directions, i.e., posing various instructions.
Experimental results show that IDEAS can generate diverse and in-depth instructions.
The constructed instructional dialogues can be used to improve the performance of the downstream chat model.
We hope IDEAS will inspire more human-inspired methods to build higher-quality instructional dialogues.

\section*{Limitations}

\paragraph{Technical Limitations.}
Due to limited computational and financial resources, we only select the fine-tuned LLM as the user simulator due to its lightweight and free.
In fact, our designed IDEAS is also applicable to user simulators implemented by role-playing LLMs.
Some researches~\cite{wang2023hypothesis,zhu2023large,sun2024itd} propose similar insights, which have confirmed that using abstract rules can improve the performance of LLMs based on instruction prompts.
In addition, we select the 13B model for fine-tuning.
Recent research suggests that LLMs exhibit emerging capabilities when they expand beyond a certain threshold~\cite{wei2022emergent}.
Using IDEAS for user simulators based on larger LLM implementation may bring different effects.
However, we do not have enough resources to conduct these relevant experiments.
We welcome further researchers to study the benefits of abstract rules, i.e., instruction strategies, to user simulators of different scales and implementations.
Besides, since this is the first study to propose an inductive-deductive strategy reuse method to propose diverse and in-depth instructions, our method design is relatively simple. From the experiment, the current IDEAS is also feasible and effective. We will optimize the design of IDEAS in the future.

\paragraph{Application Limitations.}
When using IDEAS for building instructional dialogues with a specific domain, it is necessary to ensure that there are a certain number of human-machine dialogues in the relevant domain to extract the domain-dependent general strategies. This is also consistent with people's cognition that it still takes a certain amount of experience to abstract general guiding rules in a specific domain.
When adapting high-level strategies to specific domains, our designed ranker can select the set of appropriate strategies. These strategies may be domain-independent general strategy or domain-dependent general strategy. The domain-independent general strategy means that the instruction can be adapted in most domains, e.g.,``eliciting targeted information by asking context-specific questions to deepen understanding, expand knowledge, or facilitate decision-making''. The domain-dependent general strategy means that the strategy may only adapted to certain specific domains. For example, the strategy ``emotional elicitation and empathetic inquiry'' is more suitable for the domains ``role-playing'' and ``chitchat''.

This paper chooses to extract strategies from ShareGPT data and selects opening lines from the domains covered by ShareGPT for testing.
The top 10 most frequent strategies extracted from ShareGPT are shown in Table~\ref{tab:top_strategy}.
We further conduct the following experiment, i.e., manually select 8 common domains in instructional dialogues to analyze the domain distribution of instructions in ShareGPT data and our constructed dialogue datasets. The distributions are shown in Table~\ref{tab:extract_distribution} and~\ref{tab:test_distribution}.
We observe that ShareGPT contains almost no dialogues from other domains.
All opening lines we use come from these 8 common domains.
Besides, the domain distribution of our generated instructions in the ShareGPT scenario is very similar to that of the ShareGPT dataset. Further, we find that our generated instructions in the UltraChat scenario mainly cover two domains: theoretical knowledge and applied knowledge. This is mainly because UltraChat contains plenty of knowledge-based instructional dialogues.
We also find that ShareGPT also includes the chitchat dialogues, not just task-completing instructions. Chitchat can actually be regarded as a special kind of instruction since chitchat-dialogue generation can also introduce instruction strategies to guide.

\paragraph{Reproducibility of Closed Access Models.}
We implement instruction strategy extraction and generalization, along with quality control by prompting GPT-4.
We also choose GPT-4 as the system agent.
However, GPT-4 is only accessed through an interface. 
The mechanisms behind these interfaces may change at any time, so the constructed dialogues from different periods may change. 

\section*{Ethics Statement}

Given that GPT-4 is trained on online data, it is conceivable that GPT-4 may encode pervasive biases that perpetuate stereotypes, discrimination, or marginalization of specific languages or communities.
This results in potentially toxic and harmful answers from GPT-4 as the system agent.
In addition, our evaluation process involves manual intervention by two professional annotators. Each annotator is compensated $\$0.2$ per instance.

\bibliography{anthology,custom}
\bibliographystyle{acl_natbib}

\clearpage

\appendix

\begin{table}[t!]
\centering
\small
\begin{tabular}{lcc}
\toprule
\textbf{Benchmark} & \multicolumn{1}{l}{\textbf{Avg. \# Turns}} & \multicolumn{1}{l}{\textbf{Total \# Dialogues}}  \\
\toprule
AlpacaEval & 1                                 & 805                                                                  \\
MT-Bench   & 2                                 & 80                                                                   \\
MT-Bench++ & 8                                 & 80                                                                  \\
MT-Eval    & 6.96                              & 168       \\
\bottomrule
\end{tabular}
\caption{Statistics of the used benchmarks.}
\label{tab:statistics}
\end{table}
\section{IDEAS}
\label{sec:appendix}

\begin{table*}[]
\centering
\small
\resizebox{\textwidth}{!}{
\begin{tabular}{lccccccccc}
\toprule
         & \textbf{Coding}  & \begin{tabular}[c]{@{}l@{}}\textbf{Applied}\\ \textbf{Knowledge}\end{tabular} & \begin{tabular}[c]{@{}l@{}}\textbf{Role} \\ \textbf{Playing}\end{tabular} & \begin{tabular}[c]{@{}l@{}}\textbf{Theoretical} \\ \textbf{Knowledge}\end{tabular} & \textbf{Brainstorm} & \textbf{Reasoning} & \textbf{Chitchat} & \textbf{Math} & \textbf{Others}   \\ \toprule
ShareGPT & 34.10\% & 27.99\%                                                     & 14.01\%                                                 & 7.13\%                                                           & 6.73\%        & 5.23\%    & 3.95\%  & 0.77\%  & 0.09\%\\ \bottomrule
\end{tabular}
}
\caption{The domain distribution of ShareGPT data that are used to extract instruction strategies.}
\label{tab:extract_distribution}
\end{table*}

\begin{table*}[]
\centering
\small
\resizebox{\textwidth}{!}{
\begin{tabular}{lccccccccc}
\toprule
         & \textbf{Coding}  & \begin{tabular}[c]{@{}l@{}}\textbf{Applied}\\ \textbf{Knowledge}\end{tabular} & \begin{tabular}[c]{@{}l@{}}\textbf{Role} \\ \textbf{Playing}\end{tabular} & \begin{tabular}[c]{@{}l@{}}\textbf{Theoretical} \\ \textbf{Knowledge}\end{tabular} & \textbf{Brainstorm} & \textbf{Reasoning} & \textbf{Chitchat} & \textbf{Math} & \textbf{Others}   \\ \toprule
ShareGPT & 27.84\% & 23.06\%                                                     & 18.51\%                                                 & 14.85\%                                                           & 3.70\%        & 4.09\%    & 4.57\%  & 1.31\% & 0.00\% \\ 
UltraChat & 0.73\% & 38.72\%                                                     & 2.96\%                                                 & 50.28\%                                                           & 1.13\%        & 5.04\%    & 1.10\%  & 0.03\% & 0.00\% \\ 
\bottomrule
\end{tabular}
}
\caption{The domain distribution of opening lines in ShareGPT and UltraChat dialogue scenarios.}
\label{tab:test_distribution}
\end{table*}

\subsection{Model Training}
\paragraph{Ranker.}
\label{appendix:ranker}
The BERT-based ranker may have limited performance.
Therefore, this may raise a question: why not train a stronger ranker to select the top-$1$ strategy or randomly select a strategy from the top-$k$ strategies, i.e., the candidate, to avoid further selection by the user simulator?
This is mainly due to the following reasons:
(1) There are generally multiple appropriate instruction strategies for a given dialogue history, so it is unreasonable to choose the top-$1$ strategy.
(2) It is reasonable for the user simulator to sample one from multiple appropriate strategies in the candidate.
(3) Limited performance means that not every strategy in the candidate is appropriate, so random selection from the candidate is also unreasonable.
(4) The strategy selection of the user simulator is better than that of the ranker, whose secondary selection can select more appropriate strategies.
(5) It is easy for the user simulator to select a strategy before generating instructions, but training a stronger ranker will greatly increase the complexity of IDEAS.
Overall, this lightweight model is sufficient for a coarse selection of candidates.

\section{Experimental Details}
\label{appendix:details}
\subsection{Data}
The basic data used for all experiments are collected by~\citet{sun2023parrot}.
For instruction generation, the human-machine instructional dialogues comes from ShareGPT~\footnote{https://huggingface.co/datasets/shareAI/ShareGPT-Chinese-English-90k}~\footnote{https://huggingface.co/datasets/Aeala/ShareGPT\_Vicuna\\\_unfiltered} to obtain instruction strategies, along with training the user simulator and the ranker in the induction stage.
The dialogues are further cleaned by using the provided method~\footnote{https://huggingface.co/datasets/shareAI/ShareGPT-Chinese-English-90k/blob/main/shareGPT/filter\_data.py}.
Finally, we obtain $56,929$ dialogues.
Subsequently, we process each multi-turn dialogue $\{\boldsymbol{q}_0, \boldsymbol{a}_0, \dots, \boldsymbol{q}_T, \boldsymbol{a}_T\}$ into $T$ pairs of dialogue histories and instructions, e.g., $<\{\boldsymbol{q}_0, \boldsymbol{a}_0, \dots, \boldsymbol{a}_{t-1}\}, \boldsymbol{q}_t>$.
We further obtain $211,495$ pairs and the corresponding original instruction strategies. 
Afterward, we achieve $1,593$ high-level instruction strategies via instruction strategy abstraction.
We count the top 10 most frequent instruction strategies extracted from ShareGPT data and give the number and ratio of occurrences in Table~\ref{tab:top_strategy}.
Furthermore, we randomly sample $200$ pairs of dialogue histories and instructions and employ three human annotators to evaluate the appropriateness of original and high-level instruction strategies.
About $95.67\%$ original instruction strategies are accepted by at least two annotators, and about $95\%$ high-level instruction strategies are accepted.
Similarly, we evaluate the accuracy of the original strategy abstraction, and at least two annotators accept about $94.33\%$ corresponding high-level strategies.

In the deduction stage, we randomly sample $10,000$ dialogues from unused ShareGPT and UltraChat datasets~\cite{sun2023parrot} respectively, and then intercept the first-turn dialogues $\{\boldsymbol{q}'_0, \boldsymbol{a}’_0\}$ as opening lines.
We set the maximum interaction to $10$ rounds and the maximum of regeneration in quality control to $5$.
The average number of regeneration is $0.69$.
We also evaluate the proportion of instruction strategies selected by the user simulator from the candidates, which is $97.76\%$.
Besides,  we evaluate the appropriateness of the selected strategies by the user simulator, and the annotators accept about $93.82\%$ selected instruction strategies.
Furthermore, we manually evaluate the correctness of quality control, and at least two annotators accepted $93.33\%$ of the judgment results.
After achieving our constructed instructional dialogues $\mathcal{D}_{chat}$, we clean dialogues from $\mathcal{D}_{chat}$ by removing subsequent rounds starting from the round in which the error content appears.
The error content contains the empty instructions and the incorrect answers caused by exceptions in calling the interface API of the system agent.

\subsection{Implementation Details}
\label{appendix:detail}
\paragraph{IDEAS.}
To validate the effectiveness of IDEAS, it is necessary to ensure the quality of each module when implementing IDEAS. To this end, we choose the top-tier model currently, i.e., GPT-4.
Therefore, we prompt \texttt{gpt-4-1106-preview} for instruction strategy extraction and abstraction, along with quality control.
We set the temperature to $0$, the presence\_penalty to $0.6$, and the frequency\_penalty to $0$.
For strategy clustering, the threshold $\epsilon$ is set to $0.5$.
The entire cluster process is implemented by Faiss vectorstore~\cite{johnson2019billion}. 
For ranking, we set the threshold $\eta$ to $0.5$.
We limit the size of the candidates to 50 due to the context length limitation.
When generating instructions, we use FasterTransformer~\footnote{https://github.com/NVIDIA/FasterTransformer} for inference acceleration.
And the temperature is set to $0.7$, the top-p is $0.9$, and the max new tokens is $96$.
The average inference time for one instruction is approximately $0.2$ second.
Besides, we follow existing work and employ the top-tier model, i.e. \texttt{gpt-4-1106-preview} as the system agent, where the parameters are default.

\paragraph{User Simulator.}
The user simulator is built by fine-tuning LLaMA2-13B~\cite{touvron2023llama} for $3$ epochs, with the learning rate of 2e-5, the batch size of $1024$, and the max model length of $4096$. 
We train the user simulator on $64$ A100-80G GPUs for approximately $5$ hours. 
We adopt the last checkpoint for instruction generation.

\paragraph{Ranker.}
The ranker is built by fine-tuning BERT-base~\cite{devlin-etal-2019-bert} for $10$ epochs, with the learning rate of 2e-5, the batch size of $2048$, and the max sequence length of $512$.
We train the ranker on $8$ A100-80G GPUs for approximately $4.5$ hours.
We finally adopt the last checkpoint for ranking.

\paragraph{Chat Models.}
The chat models are also built by fine-tuning LLaMA2-13B on the different instructional dialogues constructed by IDEAS and other baseline methods in Section~\ref{setup:baseline} respectively.
Following Vicuna's training schema~\cite{vicuna2023}, we train chat models by learning the output of the system agent and ignoring the user instructions.
We train chat models on $1$ A100-80G GPUs for $3$ epochs, with the learning rate of 2e-5, the batch size of $256$, and the new model length of $4096$.
The training lasts $2.5$ hours.
We finally adopt the last checkpoint for evaluation.

\begin{table*}[t]
\centering
\resizebox{\textwidth}{!}{
\begin{tabularx}{\textwidth}
{>{\raggedright\arraybackslash}p{1cm} >{\raggedright\arraybackslash}p{10cm}>{\raggedright\arraybackslash}p{2cm} >{\raggedright\arraybackslash}p{2cm}}
    \toprule
    \textbf{Rank}    & \textbf{High-level Instruction Strategy} & \textbf{Percentage} & \textbf{Count}\\
    \midrule
    1     & Eliciting targeted information by asking context-specific questions to deepen understanding, expand knowledge, or facilitate decision-making     &$3.95\%$     & 8337     \\
    \midrule
    2     & Requesting specific, practical examples and real-world applications to clarify and deepen understanding of theoretical concepts, ensure comprehensive comprehension, and facilitate the application of knowledge in various contexts     &$3.92\%$     & 8273     \\
    \midrule
    3     & Requesting elaboration, clarification, and additional information to enhance understanding and obtain comprehensive context     &$3.32\%$     & 7007   \\
    \midrule
    4     & Requesting detailed, step-by-step information or guidance tailored to specific tasks, scenarios, or contexts to ensure accurate understanding and implementation     &$2.72\%$     & 5740     \\
    \midrule
    5     & Seeking practical and actionable solutions through clarification, troubleshooting, and expert guidance for identified problems within various contexts     &$2.57\%$     & 5424     \\
    \midrule
    6     & Clarifying Concepts and Terminology     &$2.22\%$     & 4685     \\
    \midrule
    7     & Encourage detailed elaboration and comprehensive understanding through open-ended, context-specific inquiry     &$1.92\%$     & 4052     \\
    \midrule
    8     & Strategic Inquiry for Expert Insight on Complex Topics     &$1.42\%$     & 2997     \\
    \midrule
    9     & Proposing and soliciting feedback for enhancements, alternatives, and improvements to existing solutions, methods, or ideas     &$1.42\%$     & 2997   \\
    \midrule
    10     & Applying and adapting knowledge and concepts to diverse contexts and scenarios for enhanced understanding and problem-solving     &$1.37\%$     & 2891   \\
    
    \bottomrule
\end{tabularx}
}
\caption{The top 10 most frequent instruction strategies in ShareGPT data.}
\label{tab:top_strategy}
\end{table*}

\begin{table*}[t]
\centering
\resizebox{\textwidth}{!}{
\begin{tabularx}{\textwidth}
{>{\raggedright\arraybackslash}p{15.5cm}}
    \toprule
    \textbf{Prompt for Instruction Strategy Extraction} \\
    \midrule
You are very good at learning from real-human dialogues about how they generate high-quality instructions. The dialogue history is: \{Dialogue History\}. The next instruction generated by the human is: \{Instruction\}.

~

Please analyze why humans generate such instruction based on the above dialogue history, and give the instruction strategy you have learned. 

~

Good instruction strategies need to meet the following requirements: \\
(1) Instruction strategies should be expressed in simple phrases; \\
(2) Instruction strategies should not include specific information from the dialogue history. 

~

To better assist you in inducing instruction strategies, I will provide you with some examples: \\
(1) Explore possibilities and propose hypotheses; \\
(2) Provide detailed information and seek specific solutions; \\
(3) Dig deep into details. \\
These examples are used to help you clarify the expression form of instruction strategies, which are not entirely effective instruction strategies. You need to induce them yourself.

 ~
 
Finally, please output in the following format: \{``analysis'': xxx, ``strategy'': xxx\}, where ”analysis” is the reason why humans generate such instruction based on the dialogue history, and ”strategy” is just one induced instruction strategy. The expression should be concise, with no more than 20 words. Now please start thinking seriously. The results of your thinking are very important to me. \\
    \bottomrule
\end{tabularx}
}
\caption{The prompt for instruction strategy extraction.}
\label{tab:extraction_prompt}
\end{table*}

\begin{table*}[t]
\centering
\resizebox{\textwidth}{!}{
\begin{tabularx}{\textwidth}
{>{\raggedright\arraybackslash}p{15.5cm}}
    \toprule
    \textbf{Prompt for Instruction Strategy Abstraction} \\
    \midrule
You are very good at generalizing a series of original instruction strategies into a high-level instruction strategy.

~

High-level instruction strategies are general principles for thoughtfully, purposefully, and strategically generating instructions to obtain the desired information, create dialogues, stimulate thinking, or encourage others to share.        

~

To better reuse instruction strategies in new dialogue scenarios, you need to generalize a series of similar original instruction strategies into a higher-level instruction strategy. Each original instruction strategy is a specification of a high-level instruction strategy.                    

~

We provide a list of similar original instruction strategies:\{The List of Instruction Strategies\}, and ask you to generalize them into a high-level instruction strategy. The high-level instruction strategy need to be a simple phrase. Next, please directly output the high-level instruction strategy: xxx. \\
    \bottomrule
\end{tabularx}
}
\caption{The prompt for instruction strategy abstraction.}
\label{tab:abstraction_prompt}
\end{table*}

\begin{table*}[t]
\centering
\resizebox{\textwidth}{!}{
\begin{tabularx}{\textwidth}
{>{\raggedright\arraybackslash}p{16cm}}
    \toprule
    \textbf{Prompt for User Simulator} \\
    \midrule
 You are an expert at generating instructions in human-AI conversations. 
 
 ~
 
 The following is a dialogue between a user and an AI assistant. User statements start with [USER] and AI assistant statements start with [ASSISTANT]. 
 
 ~
 
 To make it easier for you to ask for an insightful instruction, I will provide you a candidate of instruction strategies. Instruction strategies are general principles for thoughtfully, purposefully, and strategically generating instructions to obtain desired information, create conversation, stimulate thinking, or encourage others to share. 
 
 ~
 
 You first need to choose an instruction strategy that fits the current flow of the dialogue from the candidate, and then generate a specific instruction based on the high-level instruction strategy and the dialogue history. \
The candidate of instruction strategies is: \{Instruction Strategies Candidate\}. The dialogue history is: \{Dialogue History\}. \\
    \bottomrule
\end{tabularx}
}
\caption{The prompt for user simulator.}
\label{tab:user_prompt}
\end{table*}

\begin{table*}[t]
\centering
\resizebox{\textwidth}{!}{
\begin{tabularx}{\textwidth}
{>{\raggedright\arraybackslash}p{16cm}}
    \toprule
    \textbf{Prompt for Quality Control} \\
    \midrule
 You are an expert in data evaluation of multi-turn instruction-following dialogues. The multi-turn instruction-following dialogue is a multi-turn dialogue process in which the USER poses an instruction and the ASSISTANT gives the corresponding answer. Currently, the USER raises a new instruction based on the instruction-following dialogue history. Your task is to check whether the proposed instruction is reasonable.

 ~
 
A reasonable instruction needs to be correct and logically coherent with the dialogue history.  

~

A correct instruction needs to meet the following two conditions: \\
(1) The raised instruction must be no contradiction with the answers from ASSISTANT, or the information obtained by further reasoning based on the answers from ASSISTANT, or the information obtained from a comprehensive understanding of the answers from ASSISTANT, or the existing instructions;  \\
(2) The raised instruction cannot be answered by any existing answers from the ASSISTANT.

 ~
 
 A logically coherent instruction needs to meet at least one of the following conditions: \\
 (1) The raised instruction may be a new request, suspicion, confirmation, or inquiry about the answers from the ASSISTANT; \\
 (2) The raised instruction may be a continued instruction about the concept or entity contained in the answers from the ASSISTANT; \\
 (3) The raised instruction may be an extension of previous instructions; \\
 (4) The raised instruction may be a new instruction about the information related to the concept or entity contained in the answers from the ASSISTANT.

 ~
 
 The dialogue history is \{Dialogue History\}, the existing instructions are \{Instruction List\}, and the currently raised instruction is \{Instruction\}. Next, please carefully analyze whether the raised instruction is reasonable. Please be sure to output according to the following JSON format: ``analysis'': ``xxx'', ``result'': ``xxx''\}, where ``analysis'' is the reason for whether it is reasonable, and ``result'' is ``yes'' or ``no''. \\
    \bottomrule
\end{tabularx}
}
\caption{The prompt of quality control.}
\label{tab:qc_prompt}
\end{table*}

\begin{table*}[t]
\centering
\resizebox{\textwidth}{!}{
\begin{tabularx}{\textwidth}
{>{\raggedright\arraybackslash}p{16cm}}
    \toprule
    \textbf{Prompt for Instruction Evaluation} \\
    \midrule
 You are a helpful, harmless, and precise assistant who checks the quality of the human’s instructions in the following multi-turn instruction dialogues.

 ~
 
 We would like to ask for your feedback on the quality of the human instructions based on the following evaluation metrics:\\
 (1) Appropriateness, which means whether the instruction is appropriate with the dialogue history and should not be detached from the context or topic. \\
 (2) Coherence, which means whether the instruction is logically coherent to the above instructions or the answers in the dialogue history. \\
 
 (3) Depth, which means whether the instruction expands the topic or explores more details in the dialogue history.\\
 (4) Insight, which means whether the instruction can bring new understanding, stimulate thought, improve the quality of the dialogue, or help unearth more valuable information, instead of repeating known content.\\
 (5) Diversity, which means whether the instruction is different from the above instructions. The difference is reflected in the instruction type. 
 The instruction type might involve data processing, fact-based inquiries, opinion-based inquiries, hypothetical scenarios, exploratory prompts, and physical actions, etc.

 ~
 
 The multi-turn instructional dialogue history is: \{Dialogue History\}. \\
 The current instruction is: \{Instruction\}.

  ~
  
 Each evaluation metric counts for 10 points and you will rate the current instruction based on this dialogue history, with a high score representing better performance. You should carefully analyze the quality, and then give each score. Please be sure to output in the following JSON format: \{``analysis'': ``[the quality in five evaluation dimensions]'', ``score'': \{``Appropriateness'': ``[an integer number between 1 and 10]'', ``Coherence'': ``[an integer number between 1 and 10]'', ``Depth'': ``[an integer number between 1 and 10]'', ``Insight'': ``[an integer number between 1 and 10]'', ``Diversity'': ``[an integer number between 1 and 10]''\}\}.\\

    \bottomrule
\end{tabularx}
}
\caption{The prompt for evaluating the generated instruction.}
\label{tab:data_eval_prompt}
\end{table*}

\begin{table*}[t]
\centering

\resizebox{\textwidth}{!}{
\begin{tabularx}{\textwidth}
{>{\raggedright\arraybackslash}p{16cm}}
    \toprule
    \textbf{A Constructed Multi-Turn Instructional Dialogue by Parrot-Ask} \\
    \midrule
 USER: Are there any major cultural events or celebrations that are specific to indigenous or Aboriginal Australians?

 ~
  
 ASSISTANT: Yes, there are several major cultural events and celebrations that are specific to indigenous or Aboriginal Australians. Here are a few examples:1. NAIDOC Week: This is a week-long celebration held every July that celebrates the history, culture, and achievements of Aboriginal and Torres Strait Islander peoples.2. Sorry Day: This is held on May 26th and is a day to remember the Stolen Generations, the children who were taken away from their families by the Australian government between 1910 and 1970.3. Corroboree: This is a traditional Aboriginal ceremony that often includes dance, song, and storytelling, and is used to pass on cultural knowledge and history.4. Dreamtime stories and festivals: These are traditional stories that explain the creation of the world and the importance of different animals and landmarks. Many festivals throughout the year celebrate these stories and the culture they represent.5. Smoking ceremonies: These are traditional cleansing ceremonies that use smoke from native plants like eucalyptus to purify people and their surroundings. They are often used in traditional Aboriginal events and ceremonies.

 USER: Are there any major sporting events or competitions that are popular in Australia? 

 ~

 ASSISTANT: Certainly, Australia is home to a vibrant sporting culture, and there are several major sporting events and competitions that are extremely popular across the country. Some of the noteworthy ones include:1. **AFL Grand Final:** The Australian Football League (AFL) Grand Final is a major Australian rules football match that traditionally takes place at the Melbourne Cricket Ground on the last Saturday in September. It determines the league's premiers.2. **The Ashes:** When played in Australia, The Ashes-a cricket test series between England and Australia-is one of the nation's most eagerly awaited sporting events.3. **Australian Open Tennis:** Held in Melbourne each January, the Australian Open is one of the four Grand Slam tennis events held globally and is the first of the year.4. **Melbourne Cup:** This is Australia's most famous annual Thoroughbred horse race. Held on the first Tuesday in November, it's a major event for horse racing enthusiasts and is known as ``the race that stops a nation.'' 5. **Sydney to Hobart Yacht Race:** Starting in Sydney on Boxing Day (December 26), the Sydney to Hobart Yacht Race is an iconic event that sees yachts race from Sydney Harbour to Hobart, Tasmania.6. **National Rugby League (NRL) Grand Final:** This is the concluding playoff game of the NRL season and is played between the top teams to decide on the premiers of the league.7. **Bledisloe Cup:** An international rugby union competition played between the Australian Wallabies and the New Zealand All Blacks, it's considered one of the fiercest rivalries in rugby.8. **Bathurst 1000:** A 1,000-kilometre motor racing event held annually at Mount Panorama Circuit in Bathurst, it's widely regarded as the pinnacle of motorsport in Australia.9. **Gold Coast Marathon:** This is a popular running event that attracts competitors from all over the world. It's usually held in July.10. **The Boxing Day Test Match:** Held at the Melbourne Cricket Ground, this is one of the most popular cricket events in Australia, starting annually on December 26th.These events are just a sampling of the vast array of sporting competitions that capture Australia's sporting spirit. In addition to these, there are many other national and regional competitions across sports like soccer, basketball, netball, and more that enjoy widespread popularity.

 USER: What are some major natural or man-made landmarks in Australia that are worth visiting?

 ~
 
 ASSISTANT: Australia is renowned for its diverse landscapes and iconic landmarks, both natural and man-made. Here are some notable examples:1. **Great Barrier Reef:** The world's largest coral reef system, located off the coast of Queensland. It's a world heritage site and one of the seven natural wonders of the world.2. **Uluru (Ayers Rock):** A massive sandstone monolith in the heart of the Northern Territory's Red Centre desert. It is sacred to the indigenous Anangu people and a UNESCO World Heritage Site.3. **Sydney Opera House:** An iconic, UNESCO World Heritage-listed building on Sydney Harbour, known for its distinctive sail-like design.4. **Sydney Harbour Bridge:** An iconic\\
\end{tabularx}
}
\caption*{}
\label{tab:parrot_case}
\end{table*}

\begin{table*}[t]
\vspace*{\fill}
\centering
\resizebox{\textwidth}{!}{
\begin{tabularx}{\textwidth}
{>{\raggedright\arraybackslash}p{16cm}}
 bridge in Sydney, known locally as ``the Coathanger'', which offers breathtaking views of the harbor and city.5. **The Twelve Apostles:** A series of limestone stacks off the shore of Port Campbell National Park, by the Great Ocean Road in Victoria.6. **Blue Mountains:** Named for the blue haze from the eucalyptus trees, this stunning natural region is just west of Sydney and known for its dramatic scenery.7. **Kakadu National Park:** Located in the Northern Territory, it's Australia's largest national park, with diverse wildlife and ancient Aboriginal rock art sites.8. **Bondi Beach:** Located in Sydney, this is one of the world's most famous beaches, known for its golden sands and excellent surf conditions.9. **Daintree Rainforest:** Situated in North Queensland, it's one of the oldest rainforests in the world and a part of the Wet Tropics of Queensland World Heritage Site.10. **Fraser Island:** The world's largest sand island, located off the eastern Queensland coast, is known for its unique dune formations and rainforest.11. **The Pinnacles:** Located in Nambung National Park in Western Australia, this desert area features thousands of limestone pillars scattered across a sand landscape.12. **Cradle Mountain:** Part of the Tasmanian Wilderness World Heritage Area, it provides dramatic mountainous scenery and opportunities for wildlife encounters.13. **Rottnest Island:** Off the coast of Western Australia, known for its beautiful beaches and friendly quokkas.14. **Freycinet National Park:** Home to the stunning Wineglass Bay, this national park on Tasmania's east coast showcases incredible natural beauty and diverse ecology.15. **Barrier Reef:** Not only is the Great Barrier Reef itself a major attraction, but many of the islands that dot the reef, such as Hamilton Island and the Whitsundays, are popular destinations too.These remarkable sites offer a glimpse into the natural beauty and rich cultural heritage of Australia, making them worthwhile destinations for travelers seeking both adventure and an appreciation for some of the world's most spectacular landscapes.

 USER: Are there any major natural or man-made landmarks in Australia that are worth visiting?

 ~
 
 ASSISTANT: Australia is home to many unique and spectacular natural and man-made landmarks that attract visitors from around the world. Here are some key landmarks and destinations to consider:**Natural Landmarks:**1. **Great Barrier Reef:** The world's largest coral reef system, located in the Coral Sea off the coast of Queensland, is famous for its stunning marine life and is a UNESCO World Heritage Site.2. **Uluru (Ayers Rock):** This massive red sandstone monolith in the Northern Territory is sacred to the indigenous Anangu people and is known for its changing colors at sunrise and sunset.3. **The Twelve Apostles:** These are limestone stacks off the coast of the Great Ocean Road in Victoria. They have been formed by erosion and are a popular attraction to view at sunrise or sunset.4. **Kakadu National Park:** Situated in the Northern Territory, this park features diverse ecosystems, significant Aboriginal rock art, and abundant wildlife.5. **Daintree Rainforest:** Located in northeastern Queensland, it is the oldest tropical rainforest in the world and a UNESCO World Heritage Site.6. **Fraser Island:** Off the coast of Queensland, this is the largest sand island in the world, home to unique dune landscapes, rainforests, and freshwater lakes.**Man-Made Landmarks:**1. **Sydney Opera House:** An architectural icon located on Sydney Harbour, it has a distinctive series of white shell-like structures and hosts numerous performances.2. **Sydney Harbour Bridge:** Often referred to as ``The Coathanger'', this bridge provides spectacular views of the harbor and the Opera House and offers the popular ``BridgeClimb'' experience.3. **Melbourne Cricket Ground (MCG):** This stadium is steeped in Australian sporting history and is a must-visit for cricket and Australian Rules football fans.4. **Parliament House:** In Canberra, the nation's capital, this modern structure is the meeting place of the Parliament of Australia and offers tours to visitors.5. **Port Arthur Historic Site:** Located on the Tasman Peninsula in Tasmania, this former convict settlement is an open-air museum and a World Heritage Site.6. **Q1 Tower:** Situated on the Gold Coast, it is one of the tallest residential towers in the world, with an observation deck offering panoramic views.These landmarks only scratch the surface of what Australia has to offer, as every state and territory boasts its own set of natural wonders and cultural attractions. Each destination provides a different perspective on Australia's history, culture, and ecological diversity.

 USER: What are some of the most popular beaches or beach towns in Australia?\\
\end{tabularx}
}
\caption*{}
\label{tab:parrot_case}
\end{table*}

\begin{table*}[t]
\vspace*{\fill}
\centering
\resizebox{\textwidth}{!}{
\begin{tabularx}{\textwidth}
{>{\raggedright\arraybackslash}p{16cm}}
  ASSISTANT: Australia has some of the most beautiful beaches in the world, stretching across its vast coastline. With a laid-back beach culture, many of these locales are not just natural wonders but also vibrant communities with their unique charm. Here are some of the most popular beaches and beach towns in Australia:1. **Bondi Beach, New South Wales:** Perhaps Australia's most famous beach, Bondi Beach is known for its golden sands, surf culture, and the bustling Bondi to Coogee coastal walk.2. **Byron Bay, New South Wales:** A popular beach town known for its surfing, scuba diving, and a relaxed hippie vibe, complemented by boutique shops, cafes, and a picturesque lighthouse.3. **Surfers Paradise, Queensland:** Part of the Gold Coast, Surfers Paradise is renowned for its high-rise skyline, sandy beach, nightlife, and the nearby theme parks.4. **Noosa, Queensland:** A stylish resort town with a stunning national park, beautiful beaches, and a river that's perfect for kayaking and paddleboarding.5. **Manly Beach, New South Wales:** Accessible from Sydney by ferry, Manly offers a variety of beachfront restaurants, shops, and markets, along with opportunities for surfing and snorkeling at nearby Shelly Beach.6. **Broome, Western Australia:** Famous for Cable Beach's white sands and camel rides at sunset, Broome is also the gateway to the Kimberley region.7. **Margaret River, Western Australia:** Known for its surf beaches, such as Prevelly, and as a gourmet food and wine destination amidst a backdrop of stunning coastal scenery.8. **Bells Beach, Victoria:** Legendary among surfers for hosting the annual Rip Curl Pro Surfing Competition, it's one of the beaches along the Great Ocean Road.9. **Whitehaven Beach, Queensland:** Located on the Whitsunday Island, it's famous for its crystal clear waters and pristine silica sand, often regarded as one of the best beaches in the world.10. **St Kilda Beach, Victoria:** Situated in Melbourne, this cosmopolitan beach offers a long stretch of sand, a historic pier, and a vibrant arts and cafe scene.11. **Palm Beach, New South Wales:** The setting for the TV show ``Home and Away'', this Sydney beach is known for its affluent homes, iconic lighthouse, and beautiful scenery.12. **Cottesloe Beach, Western Australia:** A popular spot for swimming, snorkeling, and sunsets, located in Perth and known for its terraced lawns and Norfolk pines.13. **Port Douglas, Queensland:** A gateway to the Great Barrier Reef and Daintree Rainforest, this town has a laid-back atmosphere and scenic Four Mile Beach.14. **Coolangatta, Queensland:** The southern end of the Gold Coast, known for its relaxed atmosphere and excellent surfing spots like Snapper Rocks.15. **Lorne, Victoria:** A picturesque town on the Great Ocean Road, offering a beautiful beach, arts community, and the nearby Erskine Falls.These beaches are just a taste of the diverse beach experiences available down under, with each offering unique activities and attractions, from relaxation and natural beauty to adventure sports and vibrant nightlife.

 USER: What are the main airports in Australia?  

  ~
  
 ASSISTANT: Australia has several major airports that serve as key hubs for both international and domestic travel. Here are some of the main airports in the country:1. **Sydney Airport (SYD) - Sydney Kingsford Smith Airport:** Located in Sydney, New South Wales, it is the busiest airport in Australia and serves as a major gateway to the country.2. **Melbourne Airport (MEL) - Tullamarine:** It is the second busiest airport, located in Melbourne, Victoria, and serves as a major hub for domestic and international flights.3. **Brisbane Airport (BNE):** The main airport serving Brisbane and South East Queensland, it handles a significant amount of domestic and international traffic.4. **Perth Airport (PER):** This is the major airport in Western Australia, located near Perth, and serves as an important link between Australia and countries in Africa, Asia, and Europe.5. **Adelaide Airport (ADL):** Serving Adelaide, South Australia, this airport is a key domestic and international transport hub for the southern part of Australia.6. **Gold Coast Airport (OOL):** Located at Coolangatta, near the border of Queensland and New South Wales, it serves a region that includes the Gold Coast and northern parts of New South Wales.7. **Cairns Airport (CNS):** This airport in Queensland is a significant gateway to the Great Barrier Reef and is an important tourist destination airport.8. **Canberra Airport (CBR):** The principal airport serving Australia's capital, Canberra, it handles mainly domestic flights but has capabilities for international services.9. **Hobart International Airport (HBA):** While primarily serving domestic flights, it is the main airport for the state of Tasmania and can also accommodate international traffic.10. **Darwin \\
\end{tabularx}
}
\caption*{}
\label{tab:parrot_case} 
\end{table*}

\begin{table*}[t]
\vspace*{\fill}
\centering
\resizebox{\textwidth}{!}{
\begin{tabularx}{\textwidth}
{>{\raggedright\arraybackslash}p{16cm}}
  International Airport (DRW):** Located in the Northern Territory, this airport is a key link to Asia and services both domestic and international flights.These airports are well-connected with other major cities around the world and are equipped with a range of passenger services and facilities to ensure smooth transit for travelers.

 USER: How long does it take (in flights and layovers) to get from Sydney to ... 

 ~
 
 ASSISTANT: I'm sorry, it looks like your question might have been incomplete. If you let me know the specific destination you're interested in, I can provide you with information on the typical duration of flights and layovers from Sydney. Flight times can vary significantly depending on the final destination, the airline, the flight path, and the length and number of any layovers.

 USER: How long does it take (in flights and layovers) to get from Sydney to Los Angeles?

 ~
 
 ASSISTANT: The flight duration from Sydney (SYD) to Los Angeles (LAX) typically falls within the range of 13 to 15 hours for a non-stop flight. However, the actual duration may vary due to factors such as wind speeds, changes in flight paths, and the specific aircraft being used. Non-stop flights are available from multiple airlines and are the quickest option for traveling between these two cities.If your flight includes a layover, the total travel time will be longer and will depend on the duration of the layover. For example, if you have a layover in cities like Auckland, Honolulu, or Dallas, the layover time can range from approximately one hour to several hours, which will be added to the overall travel time. In the case of layovers, it's not uncommon for the total journey to take anywhere from 16 to 24 hours or even longer, particularly if the layover is extended.It is always advisable to consult the airlines for the most up-to-date flight schedules and durations, as they can be subject to changes based on factors such as seasonal variations, demand, and operational considerations.

 USER: What are the main tourist destinations in Australia?

 ~
 
 ASSISTANT: Australia is a vast country with a wide array of tourist destinations ranging from natural wonders and beautiful beaches to vibrant cities and unique wildlife. Here's a list of some main tourist destinations in Australia:1. **Great Barrier Reef:** One of the seven wonders of the natural world, located in Queensland. It's the earth's largest coral reef system and a popular spot for snorkeling and scuba diving.2. **Sydney:** Known for its Sydney Opera House, Harbour Bridge, Bondi Beach, and the historic Rocks district. Sydney is a vibrant city with a beautiful harbor and numerous attractions.3. **Uluru-Kata Tjuta National Park:** Home to the iconic Uluru (Ayers Rock) and the dome-shaped rocks known as Kata Tjuta (The Olgas), this area is rich in Aboriginal culture and offers stunning landscapes in the Northern Territory.4. **Melbourne:** Australia's second-largest city, known for its culture, arts, dining, and shopping. It's also famous for its coffee culture and hosts events like the Australian Open.5. **Gold Coast:** Renowned for its long sandy beaches, surfing spots, theme parks like Warner Bros. Movie World, Sea World, and energetic nightlife.6. **The Great Ocean Road:** A scenic drive along the Victorian coast, known for the Twelve Apostles, beautiful coastlines, and rainforest walks.7. **Kakadu National Park:** The largest national park in Australia, located in the Northern Territory, is known for its biodiversity, wetlands, and ancient Aboriginal rock art galleries.8. **Brisbane:** The capital of Queensland, known for its vibrant urban precincts, the riverside South Bank, and proximity to attractions like the Sunshine Coast and Moreton Island.9. **Tasmania:** The island state south of the Australian mainland, known for its rugged wilderness areas, historical sites such as Port Arthur, and the Museum of Old and New Art (MONA) in Hobart.10. **Perth:** The capital of Western Australia, known for its sunny weather, beautiful beaches, and Kings Park. It is also a gateway to the natural beauty of the state's south and the Ningaloo Reef in the north.11. **Daintree Rainforest:** Located in North Queensland, it's the oldest surviving tropical rainforest on Earth and sits adjacent to the Great Barrier Reef.12. **Byron Bay:** A coastal town in New South Wales, known for its surfing beaches, whale watching, and laid-back lifestyle.13. **Fraser Island:** The world's largest sand island, located off the southeastern coast of Queensland, known for its freshwater lakes, rainforest, and wildlife.14. **Adelaide:** Known as the City\\
\end{tabularx}
}
\caption*{}
\label{tab:parrot_case}
\end{table*}

\begin{table*}[t]
\vspace*{\fill}
\centering
\resizebox{\textwidth}{!}{
\begin{tabularx}{\textwidth}
{>{\raggedright\arraybackslash}p{16cm}}
  of Churches, Adelaide is the capital of South Australia and is renowned for its festivals, food, and wine, with regions like the Barossa Valley located nearby.15. **The Whitsundays:** An archipelago of 74 tropical islands off the Queensland coast, amidst the Great Barrier Reef, popular for sailing, diving, and luxury resorts.These destinations are just a snapshot of what Australia has to offer tourists. Each provides its unique experiences, from natural beauty to cosmopolitan city life, catering to a wide range of interests and activities.

 USER: How long does it take (in flights and layovers) to get from Sydney to Auckland?

 ~

 ASSISTANT: The flight duration from Sydney, Australia (SYD) to Auckland, New Zealand (AKL) typically takes about 3 hours for a direct (non-stop) flight. However, the overall travel time can vary based on various factors, including the airline chosen, the flight path, and weather conditions. It is important to note that if your flight includes layovers, the total travel time could be significantly longer.Layovers commonly occur when traveling with airlines that do not offer direct service between Sydney and Auckland or when opting for more cost-effective but less direct routes. The duration of layovers can vary greatly, ranging from a brief hour or two to several hours, depending on the specific layover location and the flight schedule.To minimize your travel time, it is always advisable to check with the airline when booking your flight for the most up-to-date information on direct flight options. Direct flights are typically the fastest way to reach Auckland from Sydney. In fact, the Sydney to Auckland route is considered one of the shorter international trips available from Australia.Keep in mind that these travel times are approximate and can be subject to change. Factors such as air traffic, airport procedures, and unforeseen circumstances can affect the actual duration of your journey.\\
 
    \bottomrule
\end{tabularx}
}
\caption{A constructed multi-turn instructional dialogue by Parrot-Ask.}
\label{tab:parrot_case}
\end{table*}

\begin{table*}[t]
\centering
\resizebox{\textwidth}{!}{
    \begin{tabularx}{\textwidth}
    {>{\raggedright\arraybackslash}p{16cm}}
        \toprule
        \textbf{A Constructed Multi-Turn Instructional Dialogue by IDEAS} \\
        \midrule

 USER: Are there any major cultural events or celebrations that are specific to indigenous or Aboriginal Australians?

 ~
 
 ASSISTANT: Yes, there are several major cultural events and celebrations that are specific to indigenous or Aboriginal Australians. Here are a few examples:1. NAIDOC Week: This is a week-long celebration held every July that celebrates the history, culture, and achievements of Aboriginal and Torres Strait Islander peoples.2. Sorry Day: This is held on May 26th and is a day to remember the Stolen Generations, the children who were taken away from their families by the Australian government between 1910 and 1970.3. Corroboree: This is a traditional Aboriginal ceremony that often includes dance, song, and storytelling, and is used to pass on cultural knowledge and history.4. Dreamtime stories and festivals: These are traditional stories that explain the creation of the world and the importance of different animals and landmarks. Many festivals throughout the year celebrate these stories and the culture they represent.5. Smoking ceremonies: These are traditional cleansing ceremonies that use smoke from native plants like eucalyptus to purify people and their surroundings. They are often used in traditional Aboriginal events and ceremonies.

 USER: \textbf{[}\textbf{instruction strategy}\textbf{]} Ask for specific information to clarify details \textbf{[}\textbf{instruction}\textbf{]} Tell me more about Corroboree

 ~
 
 ASSISTANT: The term ``corroboree'' is derived from the Aboriginal word ``caribberie'', which was first used by the European settlers to describe Aboriginal ceremonies that involved singing and dancing. It encompasses a wide range of Aboriginal cultural events, including ceremonies, dances, and gatherings, which can be either sacred and private or public and celebratory.Sacred corroborees are deeply rooted in the Dreamtime - the Aboriginal understanding of the world, of its creation, and the great ancestral beings who roamed the Earth and created the landscapes, plants, and animals. These ceremonies are integral to the cultural life of the community, involving rituals and stories that are passed down through generations and are not shared with outsiders.In contrast, public corroborees can be shared with non-Aboriginal people and often serve a dual purpose: They celebrate Aboriginal culture and educate others about their customs, history, and connection to the land. These events can include traditional music, dance, body painting, and costume, which often have significant meanings related to the local Aboriginal mythology, history, or the environment.Dances in a corroboree may imitate animals, reflect hunting techniques, or retell important stories or historical events. Participants often adorn their bodies with paints derived from natural materials such as ochre, and wear traditional ornaments and headgear. These adornments hold cultural meanings specific to the clan or region.Corroborees play a crucial role in preserving Aboriginal languages and traditions, which have been passed down orally over tens of thousands of years. They are a powerful expression of Aboriginal identity and community cohesion, serving both to educate the younger generation and to affirm the resilience of Indigenous culture against a backdrop of historical and ongoing challenges.While corroborees vary significantly between different Aboriginal nations (there are over 500 different clan groups or 'nations' in Australia, each with their own culture, language, and traditions), they all hold deep cultural significance and are respected events that offer a vibrant connection to an ancient heritage that predates European settlement by millennia.

 USER: \textbf{[}\textbf{instruction strategy}\textbf{]} Request detailed, step-by-step information or guidance \textbf{[}\textbf{instruction}\textbf{]} How do I find a corroboree to attend as a tourist?

 ~
 
 ASSISTANT: Attending a corroboree as a tourist can be a wonderful experience and a unique opportunity to learn more about Indigenous Australian culture. However, it's important to remember that sacred corroborees are typically closed to non-Indigenous people, as they are deeply spiritual events. If you are interested in experiencing a corroboree, you will need to look for public events that welcome visitors. \\
\end{tabularx}
}
\caption*{}
\label{tab:ideas_case}
\end{table*}

\begin{table*}[t]
\centering
\resizebox{\textwidth}{!}{
    \begin{tabularx}{\textwidth}
    {>{\raggedright\arraybackslash}p{16cm}}
  Here are a few guidelines on how to find such events:1. **Cultural Festivals and Events**: Look for cultural festivals and events in the area where you'll be traveling. These often include public corroborees or dance performances. NAIDOC Week celebrations, for example, may include events open to the public.2. **Tourist Information Centers**: Local tourist information centers can be a valuable resource. They can inform you about cultural events during your visit and whether any public corroborees are scheduled.3. **Indigenous Tourism Operators**: Consider looking for Indigenous-owned and operated tourism companies. These operators offer cultural tours and experiences that may include dance performances or opportunities to engage with other aspects of Indigenous culture.4. **Museums and Cultural Centers**: Institutions such as the National Museum of Australia, the Australian Museum in Sydney, and many others across the country often host events that showcase Aboriginal culture, including dance performances or corroboree-like presentations.5. **Online Resources**: Websites and social media pages dedicated to promoting Indigenous events can be useful for finding upcoming cultural displays, performances, and festivals that might include corroborees. 6. **Ethical Considerations**: Always approach these cultural experiences with respect. Follow protocols, ask before taking photos (if allowed at all), and listen carefully to the stories and explanations offered by Indigenous presenters. Being a considerate and respectful guest will help ensure that these important cultural traditions can continue to be shared.Remember that schedules may vary, and events may not always coincide with your visit. It's essential to plan ahead and research before your trip to maximize your opportunities for participating in these rich cultural experiences.

 USER: \textbf{[}\textbf{instruction strategy}\textbf{]} Request supplementary information and resources to enhance understanding and comprehension\textbf{[}\textbf{instruction}\textbf{]} What are some good online resources to find out about Aboriginal cultural events?

 ~
 
 ASSISTANT: There are several online resources where you can find information about Aboriginal cultural events. These websites often list events, cultural festivals, and activities that celebrate the heritage and traditions of the First Australians. Here are some places to start: 1. **NAIDOC Week Official Website**: Visit the official NAIDOC Week website (www.naidoc.org.au) to find out about events celebrating the history, culture, and achievements of Aboriginal and Torres Strait Islander peoples. NAIDOC Week is held annually in July.2. **Tourism Australia**: Check out the Tourism Australia website (www.australia.com) for information on Indigenous experiences available to visitors. They often highlight events, activities, and tours that include Aboriginal culture. 3. **Aboriginal Tourism Association**: Many regions have their own Aboriginal tourism association websites that provide information on local Indigenous culture, tours, and events.4. **State and Territory Tourism Websites**: Each Australian state and territory typically has its own tourism website with a section on Aboriginal experiences where you can learn about cultural events happening in that specific area.5. **Cultural Centres and Museums**: Visiting the websites of cultural centres and museums such as the National Museum of Australia in Canberra (www.nma.gov.au) and the Tandanya National Aboriginal Cultural Institute in Adelaide (www.tandanya.com.au) can provide information on exhibitions and events.6. **Social Media**: Follow relevant social media pages for Indigenous organizations and cultural groups, as they often promote events and share updates about cultural celebrations.7. **Eventbrite and Similar Platforms**: Websites like Eventbrite or local event listing pages may also list Aboriginal cultural events that you can participate in as a visitor.8. **Local Government Websites**: City or regional government websites are also a good resource as they may list community events, including those celebrating Aboriginal culture.

 USER: \textbf{[}\textbf{instruction strategy}\textbf{]} Request detailed, structured, and clear information in a concise and comprehensible format [\textbf{instruction}] Can you give me a list of the Aboriginal tourism associations for each region?

  ~

 ASSISTANT: Below is a list of Aboriginal tourism associations or organizations that can provide information on Indigenous cultural experiences and events in various regions of Australia. Please note, the availability and activity level of these associations may vary, and it's best to visit their respective\\
\end{tabularx}
}
\caption*{}
\label{tab:ideas_case}
\end{table*}

\begin{table*}[t]
\centering
\resizebox{\textwidth}{!}{
    \begin{tabularx}{\textwidth}
    {>{\raggedright\arraybackslash}p{16cm}}
    websites or contact them directly for the most up-to-date information:1. **Aboriginal Tourism Association of British Columbia (ATBC), Canada**   - Website: www.aboriginalbc.com (Note: This is for British Columbia in Canada, which has a significant Indigenous tourism sector, included here for broader context.)For Australian Indigenous tourism, here's a rundown by states and territories:2. **New South Wales**   - Aboriginal Tourism Association - This particular association may not have a dedicated website, but Destination NSW's website (www.visitnsw.com) often provides information on Aboriginal experiences.3. **Queensland**   - Indigenous Tourism Champions Program - Facilitated by Tourism and Events Queensland (www.teq.queensland.com)4. **Northern Territory**   - Tourism NT - While there may not be a dedicated Aboriginal tourism association, Tourism NT provides extensive information on Indigenous tourism (www.northernterritory.com)5. **Western Australia**   - WAITOC (Western Australian Indigenous Tourism Operators Council) - The leading body for Aboriginal tourism in Western Australia (www.waitoc.com)6. **South Australia**   - South Australian Tourism Commission - Often promotes Indigenous experiences throughout their website and campaigns (www.southaustralia.com)7. **Victoria**   - Aboriginal Tourism Victoria - Although not a dedicated association, Visit Victoria frequently showcases Aboriginal experiences (www.visitvictoria.com)8. **Tasmania**   - Tasmanian Aboriginal Centre - Provides information related to the Tasmanian Aboriginal community (www.tacinc.com.au)9. **Australian Capital Territory**   - VisitCanberra - The official tourism site for Canberra and the ACT, which features Aboriginal experiences (www.visitcanberra.com.au)Organizations at the national level like Indigenous Tourism Australia also play a role in promoting Aboriginal and Torres Strait Islander tourism businesses and experiences across the country.Before planning your visit or attending an event, it's always a good idea to make direct contact with the organization to obtain the most current information and ensure that what is being offered aligns with your interests and any cultural sensitivities.

 USER: \textcolor[rgb]{0, 0, 0}{\textbf{[}\textbf{instruction strategy}\textbf{]} Geographically-oriented Inquiry} \textbf{[}\textbf{instruction}\textbf{]} What are some organizations that promote Aboriginal tourism in the Northern Territory?

 ~

 ASSISTANT: Aboriginal tourism in the Northern Territory is supported by several organizations that work to promote and develop Indigenous cultural experiences for visitors. Here are some key organizations involved in this effort:1. **Northern Territory Indigenous Tourism Advisory Council (NTITAC)**   - NTITAC provides strategic advice on Indigenous tourism in the Northern Territory, aiming to support the growth and development of the sector and ensure that the cultural integrity and authenticity of Aboriginal tourism experiences are preserved.2. **Tourism NT**   - Within the Northern Territory Government, Tourism NT is responsible for promoting tourism in the region. It has a focus on Aboriginal tourism offerings through marketing initiatives and providing visitor information.3. **Parks Australia - Uluru-Kata Tjuta National Park and Kakadu National Park**   - Both parks are jointly managed by their traditional Aboriginal owners and the Australian Government. They offer unique opportunities for visitors to engage with Anangu and Bininj/Mungguy culture, respectively.4. **Indigenous Tourism Champions Program**   - While not exclusive to the Northern Territory, this program supports Aboriginal and Torres Strait Islander tourism businesses nationwide. It helps NT operators obtain recognition and promotion through Tourism Australia's marketing efforts.5. **Aboriginal Areas Protection Authority (AAPA)**   - The AAPA protects sacred sites across the Northern Territory and can provide information on appropriate behavior and protocols when visiting these areas.6. **Local Aboriginal Land Councils**   - These councils, which are specific to their regions within the Northern Territory, are often involved in tourism and can facilitate cultural experiences on Aboriginal land. They can also provide guidance and permission for accessing certain areas.7. **Local Tourism Operators**   - There are many Indigenous-owned tourism businesses in the Northern Territory that offer cultural tours, accommodation, art experiences, and more. Many of these enterprises are key promoters of Aboriginal tourism in their own rights.It's important to note that while these organizations facilitate and promote tourism, the actual cultural experiences are typically provided by local Indigenous communities, tour operators, and cultural \\
\end{tabularx}
}
\caption*{}
\label{tab:ideas_case}
\end{table*}

\begin{table*}[t]
\centering
\resizebox{\textwidth}{!}{
\begin{tabularx}{\textwidth}
{>{\raggedright\arraybackslash}p{16cm}}
 centers. When looking to participate in Aboriginal tourism experiences in the Northern Territory, it is advisable to support local and Indigenous-owned businesses, as they offer authentic and respectful insights into their culture while contributing to the local economy.

 USER: \textcolor[rgb]{0, 0, 0}{\textbf{[}\textbf{instruction strategy}\textbf{]} Elicit targeted information by asking context-specific questions} \textbf{[}\textbf{instruction}\textbf{]} What are the local land councils in the Northern Territory?

 ~
 
 ASSISTANT: Local land councils in the Northern Territory are Aboriginal organizations established under the Aboriginal Land Rights (Northern Territory) Act 1976. They represent Aboriginal landowners and residents, consult on their behalf, and negotiate in regards to the use of their land. These councils are also involved in Indigenous land management, economic development, and protection of cultural heritage. The Northern Territory has four main land councils:1. **Northern Land Council (NLC)**   - The NLC represents Aboriginal people in the Top End of the NT, including the Darwin region, Katherine area, Western Arnhem Land, and the Barkly Tablelands.2. **Central Land Council (CLC)**   - The CLC covers the southern part of the NT, including the Alice Springs region and remote communities across the central desert region.3. **Tiwi Land Council**   - This council is specific to the Tiwi Islands, which are located about 80 kilometers north of Darwin.4. **Anindilyakwa Land Council**   - This council represents the Aboriginal people of the Groote Eylandt archipelago in the Gulf of Carpentaria.Each land council has responsibilities ranging from land claims and economic development to cultural preservation and environmental protection. If you wish to visit or conduct activities on Aboriginal land in the Northern Territory, it is essential to seek the relevant land council's advice and, if required, obtain a permit. These councils not only oversee external relationships concerning land use but also play an integral part in supporting the cultural and social aspirations of Aboriginal communities.

 USER: \textcolor[rgb]{0, 0, 0}{\textbf{[}\textbf{instruction strategy}\textbf{]} Engage in Inquisitive Exploration to Deepen Understanding and Gather Comprehensive Insights} \textbf{[}\textbf{instruction}\textbf{]} What are some of the important corroborees held in the Northern Territory?
 
 ~
 
  ASSISTANT: Corroborees in the Northern Territory are cultural events that showcase the diverse traditions of the Indigenous people who live there. While some corroborees are sacred and private, others are public celebrations or festivals that may include elements of traditional dance and music that visitors can attend. Many of these events are a mix of traditional and contemporary expressions of culture. Here are some important events in the Northern Territory that tourists can often attend:1. **Barunga Festival**: Usually held in June, the Barunga Festival celebrates Aboriginal community life through music, sport, and culture, and sometimes includes ``bungal'', the local word for corroboree, with traditional dance performances.2. **Parrtjima - A Festival in Light**: Though not strictly a corroboree, Parrtjima is an annual event held in Alice Springs that showcases the oldest continuous culture on Earth through the newest technology, on a 300-million-year-old natural canvas - the MacDonnell Ranges. Some cultural talks and displays may include traditional dance performances.3. **Desert Mob**: An event held in Alice Springs, featuring a marketplace and exhibition that's one of the most significant Aboriginal art and cultural events in the nation. As part of the event, traditional dancing and storytelling might be included.4. **Merrepen Arts Festival**: The festival takes place in on the banks of the Daly River and features art, music, dance, food, and sports. Traditional dance performances are often part of the celebrations.5. **Walking with Spirits**: Held at Beswick Falls, this is a traditional corroboree hosted by the local Jawoyn people in partnership with Djilpin Arts. It's a mix of traditional and contemporary performance in a unique and remote setting, blending multimedia with traditional culture.6. **Garma Festival**: Hosted by the Yothu Yindi Foundation, Garma is held in northeast Arnhem Land. It's a premier cultural exchange event in Australia, attracting international attention. The event aims to share Yolngu culture with the world and includes traditional dance (bunggul), music, art, and cultural discussions.It's essential to be aware that dates and specifics of these events can change from year to year, so it is advisable to check the latest details when planning a visit. Also, as a sign of respect to the communities and cultures involved,\\
\end{tabularx}
}
\caption*{}
\label{tab:ideas_case}
\end{table*}

\begin{table*}[t]
\centering
\resizebox{\textwidth}{!}{
\begin{tabularx}{\textwidth}
{>{\raggedright\arraybackslash}p{16cm}}
  tourists should always adhere to the guidelines provided by the event organizers, particularly in regards to photographing or filming the proceedings.
 
 USER: \textcolor[rgb]{0, 0, 0}{\textbf{[}\textbf{instruction strategy}\textbf{]} Elicit Specific Information through Contextual and Targeted Examples} \textbf{[}\textbf{instruction}\textbf{]} What are some of the other public corroborees that tourists can attend?

 ~
 
 ASSISTANT: Throughout Australia, there are public corroborees and cultural festivals where tourists can experience Indigenous dance, music, storytelling, and art. These events are designed to be public celebrations of Indigenous culture, with many purposely created to educate and share cultural heritage with a broad audience. Here are some additional events to the ones mentioned previously:1. **NAIDOC Week**: Held across Australia in July, NAIDOC Week celebrates the history, culture, and achievements of Aboriginal and Torres Strait Islander peoples, often including public events such as marches, cultural performances, and art exhibitions.2. **Yabun Festival**: Held annually on January 26th (Australia Day) in Sydney, Yabun is the largest one-day gathering and recognition of Aboriginal and Torres Strait Islander cultures in Australia, frequently featuring live music, dance, discussions, and market stalls.3. **Tarnanthi**: An annual festival in Adelaide that showcases contemporary Aboriginal and Torres Strait Islander art. While not a corroboree, the event includes various cultural activities and may feature traditional dance and music performances.4. **The Laura Aboriginal Dance Festival**: Held in Laura, Queensland, this biennial festival is an important gathering designed to celebrate and showcase the culture of the Aboriginal communities of the Cape York Peninsula. It features dance, music, and cultural workshops.5. **Spirit Festival**: South Australia's premier Aboriginal and Torres Strait Islander arts and culture festival. It encapsulates visual art, music, dance, theater, screenings, workshops, and cultural events.6. **Homeground**: Hosted by the Sydney Opera House, this event brings together First Nations performers from Australia and around the world for a weekend of dance, music, and workshops on the Opera House's Forecourt.7. **Dance Rites**: Also hosted at the Sydney Opera House, Dance Rites is a national Indigenous dance competition that celebrates the continued cultural dance practices of First Nations communities.8. **Woodford Folk Festival**: One of the largest folk festivals in the world, this annual event in Queensland features an array of performances, including some by Aboriginal and Torres Strait Islander artists, along with activities and workshops that share Indigenous culture.Please note that the inclusion and frequency of Indigenous dances or elements resembling a corroboree can vary annually and by event. It's essential to check the latest information about each festival for the year you plan to visit and to ensure that you understand and respect the cultural protocols around photography, recording, and participation. Many of these festivals are family-friendly and are designed to foster greater understanding and appreciation of Indigenous Australian cultures.

 USER: \textcolor[rgb]{0, 0, 0}{\textbf{[}\textbf{instruction strategy}\textbf{]} Seek practical and actionable solutions through clarification or troubleshooting} \textbf{[}\textbf{instruction}\textbf{]} What are some of the challenges facing the tourism industry in the Northern Territory?

 ~
 
 ASSISTANT: The tourism industry in the Northern Territory (NT) of Australia, like many regions worldwide, faces a unique set of challenges. Some of these challenges include:1. **Remoteness and Accessibility**: The NT is a region with vast distances between major attractions and limited infrastructure, which can pose difficulties for visitors. Access to remote areas often requires 4WD vehicles and can be costly and time-consuming.2. **Climate Extremes**: The NT experiences extreme weather, including a wet season with monsoonal rains and a dry season with more moderate temperatures. This seasonality can affect access to certain areas and impact the types of activities that are available at different times of the year.3. **Sensitivity of Indigenous Sites**: Many popular tourist sites are of great cultural significance to local Indigenous communities. Managing tourist access while respecting Aboriginal cultural values and laws can be complex, requiring careful negotiations and agreements.4. **Cultural Awareness and Education**: Ensuring that tourists have an understanding and respect for local Indigenous culture is vital. There can be a challenge in balancing the education of tourists with the preservation of culture and traditions.5. **Environmental Sustainability**: The NT is home to unique\\
\end{tabularx}
}
\caption*{}
\label{tab:ideas_case}
\end{table*}

\begin{table*}[t]
\centering
\resizebox{\textwidth}{!}{
\begin{tabularx}{\textwidth}
{>{\raggedright\arraybackslash}p{16cm}}
 ecosystems and biodiversity, and there is a constant need to balance tourism development with environmental conservation. Sustainable practices are essential to protect the natural landscape for future generations.6. **Economic Viability**: The costs of operating tourism businesses in remote areas are high. This can impact the affordability and availability of tourism experiences for visitors and influence the economic sustainability for local operators.7. **Shortage of Skilled Labor**: The tourism industry often struggles with a shortage of skilled workers, particularly in remote locations, which can affect service quality and the visitor experience.8. **Competition**: The NT competes with other destinations in Australia and around the world for tourist dollars. Creating a unique value proposition for the NT is crucial in a crowded market space.9. **Marketing and Visibility**: Effectively marketing the NT in domestic and international markets is important to attract visitors. This is especially relevant as the tourism industry recovers from global disruptions like the COVID-19 pandemic.10. **Global Economic and Health Crises**: Fluctuations in the global economy, health pandemics, and other international crises can severely impact the number of tourists who visit, as seen during the COVID-19 pandemic, which resulted in border closures and a significant downturn in travel.Addressing these challenges requires cohesive efforts from the government, tourism operators, Indigenous communities, and other stakeholders. Strategies focusing on sustainable and culturally sensitive tourism, investment in infrastructure, targeted marketing, and workforce development are important components in the long-term viability and growth of tourism in the Northern Territory.\\
\bottomrule
\end{tabularx}
}
\caption{A constructed multi-turn instruction dialogue by IDEAS.
Note that the instruction strategy is only used to guide instruction generation, and the strategy is not retained in the dialogue history in practice. The expression in the table is mainly to show the strategy used to generate the specific instruction.
}
\label{tab:ideas_case}
\end{table*}

\end{document}